\documentclass[lettersize,journal]{IEEEtran}
\usepackage{amsmath,amsfonts}
\usepackage{array}
\usepackage[caption=false,font=normalsize,labelfont=sf,textfont=sf]{subfig}
\usepackage{textcomp}
\usepackage{stfloats}
\usepackage{url}
\usepackage{verbatim}
\usepackage{graphicx}
\usepackage{cite}
\usepackage{graphicx}
\usepackage{amsmath}
\usepackage{amssymb}
\usepackage{booktabs}
\usepackage{color}  

\usepackage{bm}
\usepackage{multirow}
\usepackage{bbding}
\usepackage[linesnumbered,ruled,vlined]{algorithm2e}

\begin{document}

\title{Exploiting the Partly Scratch-off Lottery Ticket for\\ Quantization-Aware Training}

\author{Yunshan Zhong,
Gongrui Nan,
Yuxin Zhang,
Fei Chao,
Rongrong Ji,~\IEEEmembership{Senior Member, IEEE}
\thanks{This work was supported by National Key R\&D Program of China (No.2022ZD0118202), the National Science Fund for Distinguished Young Scholars (No.62025603), the National Natural Science Foundation of China (No. U21B2037, No. U22B2051, No. 62176222, No. 62176223, No. 62176226, No. 62072386, No. 62072387, No. 62072389, No. 62002305 and No. 62272401), and the Natural Science Foundation of Fujian Province of China (No.2021J01002,  No.2022J06001).
}
\thanks{Y. Zhong is with Institute of Artificial Intelligence, Department of Artificial Intelligence, School of Informatics, and Key Laboratory of Multimedia Trusted Perception and Efficient Computing, Ministry of Education of China, Xiamen University, Xiamen 361005, P.R. China.}
%
%
\thanks{G. Nan, Y. Zhang, and C. Fei are with Department of Artificial Intelligence, School of Informatics, and Key Laboratory of Multimedia Trusted Perception and Efficient Computing, Ministry of Education of China, Xiamen University, Xiamen 361005, P.R. China.}
\thanks{R. Ji (Corresponding Author) is with Institute of Artificial Intelligence, and Key Laboratory of Multimedia Trusted Perception and Efficient Computing, Ministry of Education of China, Xiamen University, Xiamen 361005, P.R. China, and also with the Peng Cheng Laboratory, Shenzhen 518000, P.R. China (e-mail: rrji@xmu.edu.cn).}
\thanks{Manuscript received April 19, 2021; revised August 16, 2021.}}

\markboth{Journal of \LaTeX\ Class Files,~Vol.~14, No.~8, August~2021}%
{Shell \MakeLowercase{\textit{et al.}}: A Sample Article Using IEEEtran.cls for IEEE Journals}

\IEEEpubid{0000--0000/00\$00.00~\copyright~2021 IEEE}

\maketitle

\begin{abstract}
Quantization-aware training (QAT) receives extensive popularity as it well retains the performance of quantized networks. 
In QAT, the contemporary experience is that all quantized weights are updated for an entire training process.
In this paper, this experience is challenged based on an interesting phenomenon we observed. Specifically, a large portion of quantized weights reaches the optimal quantization level after a few training epochs, which we refer to as the partly scratch-off lottery ticket.
This straightforward-yet-valuable observation naturally inspires us to zero out gradient calculations of these weights in the remaining training period to avoid meaningless updating.
To effectively find the ticket, we develop a heuristic method, dubbed lottery ticket scratcher (LTS), which freezes a weight once the distance between the full-precision one and its quantization level is smaller than a controllable threshold.
Surprisingly, the proposed LTS typically eliminates 50\%-70\% weight updating and 25\%-35\% FLOPs of the backward pass, while still resulting on par with or even better performance than the compared baseline.
For example, compared with the baseline, LTS improves 2-bit MobileNetV2 by 5.05\%, eliminating 46\% weight updating and 23\% FLOPs of the backward pass.
Code is at \url{https://github.com/zysxmu/LTS}.

\end{abstract}


\section{Introduction}

\IEEEPARstart{D}{eep}  neural networks (DNNs) have been the foundation of many computer vision tasks for their superiority in performance. However, the success usually relies on increasing model size and computational costs, which barricades deployments of DNNs on popular resource-limited devices, such as mobile phones. In recent years, a variety of model compression techniques have been excavated to overcome this issue~\cite{han2015learning,whitepaper,hinton2015distilling,lin2020hrank}.
Among these techniques, network quantization represents the full-precision weights and activations within DNNs in a low-bit format, making it one of the most promising techniques for simultaneously reducing both the model sizes and computational costs~\cite{ACIQ,lin2020rotated,zhong2022intraq,bulat2020HighCapacity,gong2019differentiable,gao2022clusterq}.

\begin{figure*}[!t]
\centering
  \subfloat[]{
  \includegraphics[width=0.32\linewidth]{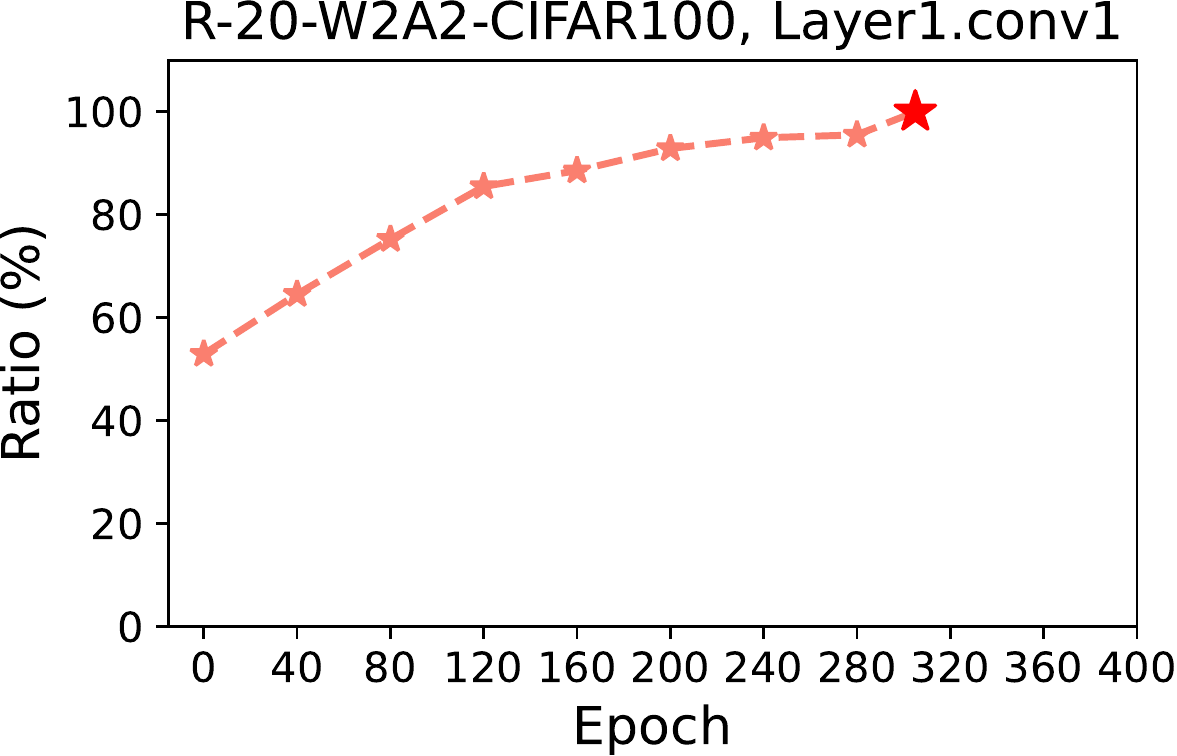}
  \label{insight:a}}
  \subfloat[]{
  \includegraphics[width=0.32\linewidth]{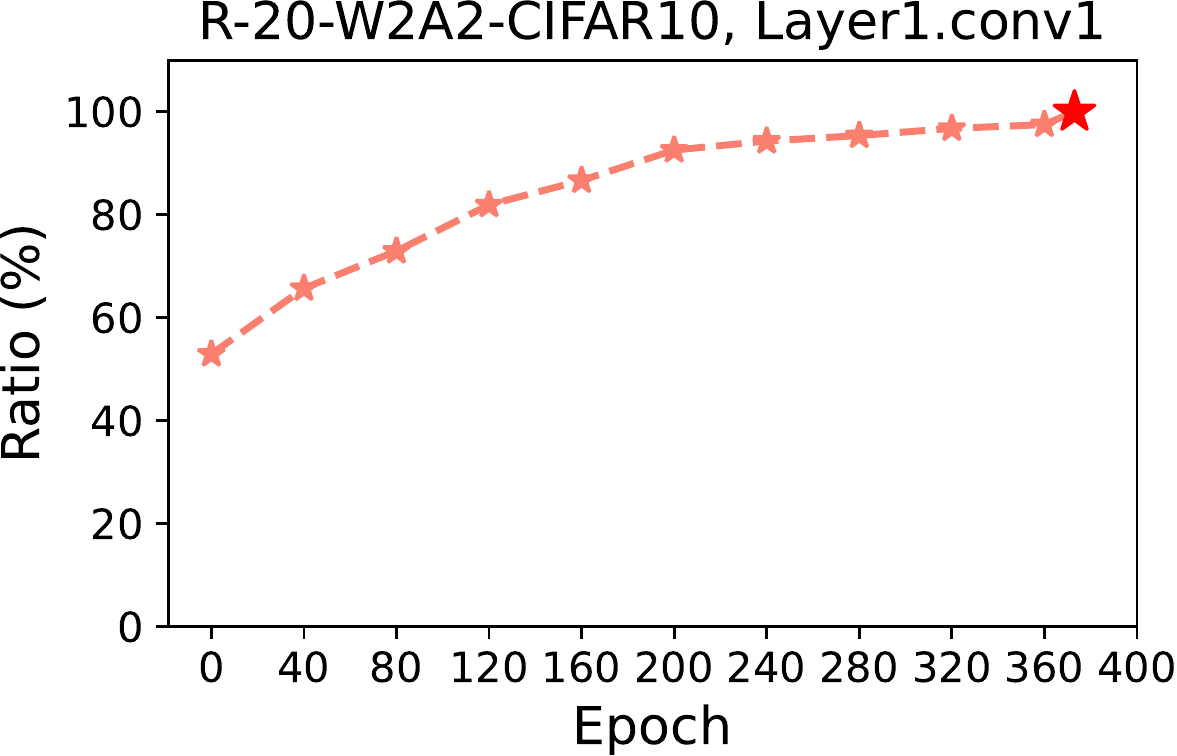}
  \label{insight:b}}
  \subfloat[]{
  \includegraphics[width=0.32\linewidth]{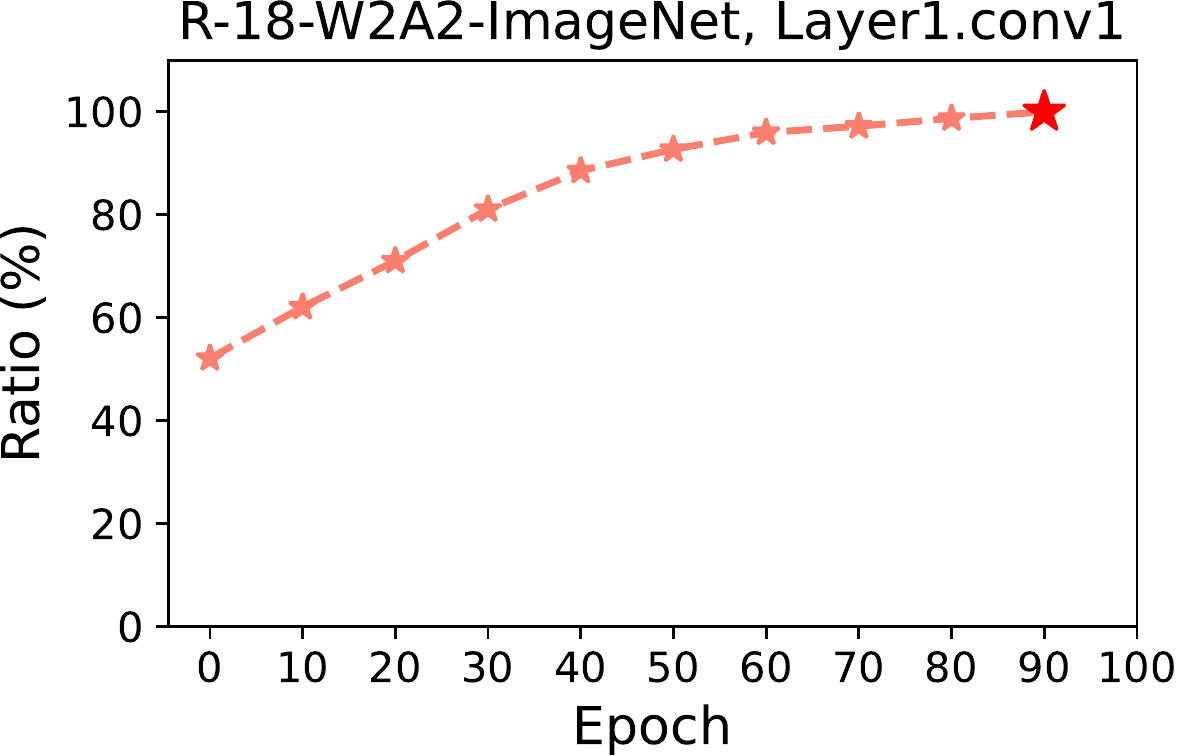}
  \label{insight:c}}
  \\
  \subfloat[]{
  \includegraphics[width=0.32\linewidth]{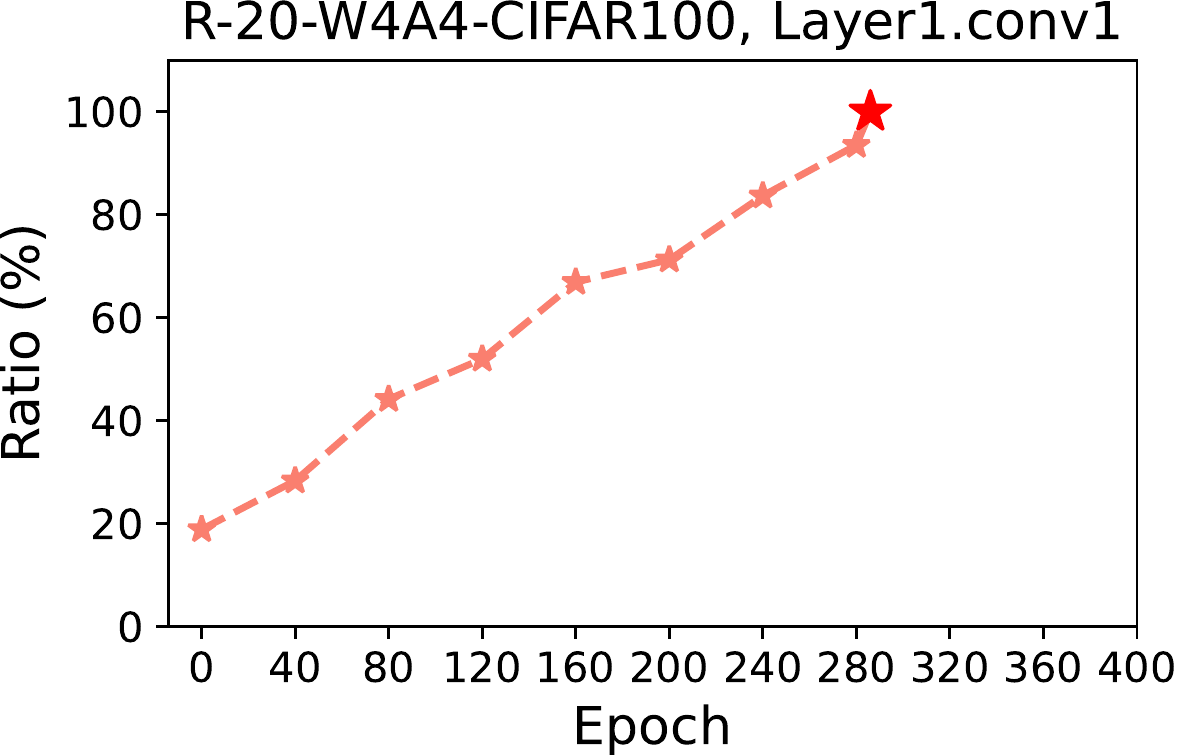}
  \label{insight:d}}
  \subfloat[]{
  \includegraphics[width=0.32\linewidth]{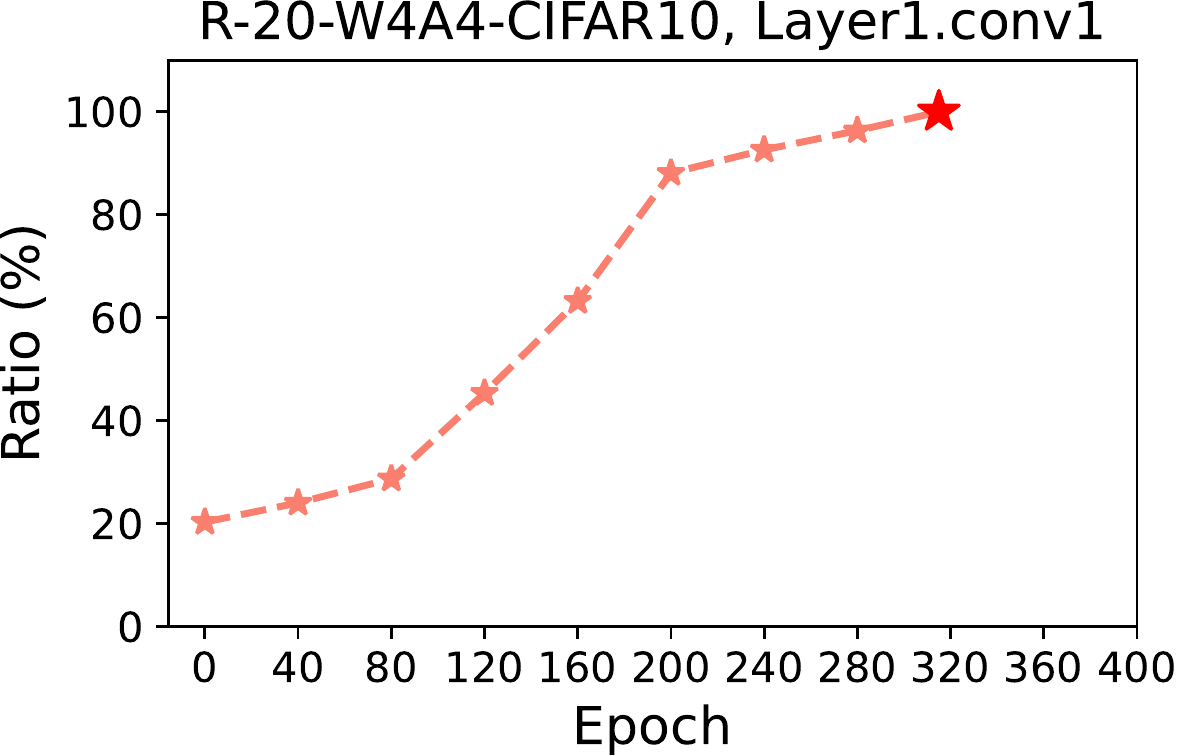}
  \label{insight:e}}
  \subfloat[]{
  \includegraphics[width=0.32\linewidth]{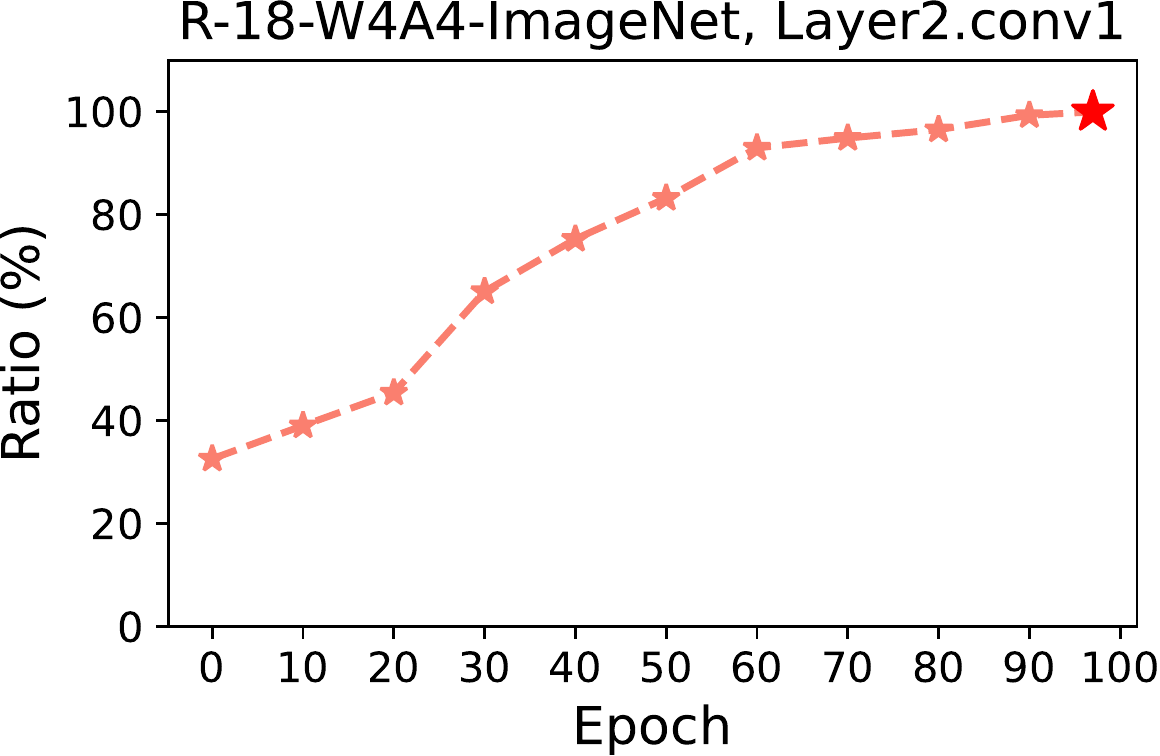}
  \label{insight:f}}
  \\
  \subfloat[]{
  \includegraphics[width=0.32\linewidth]{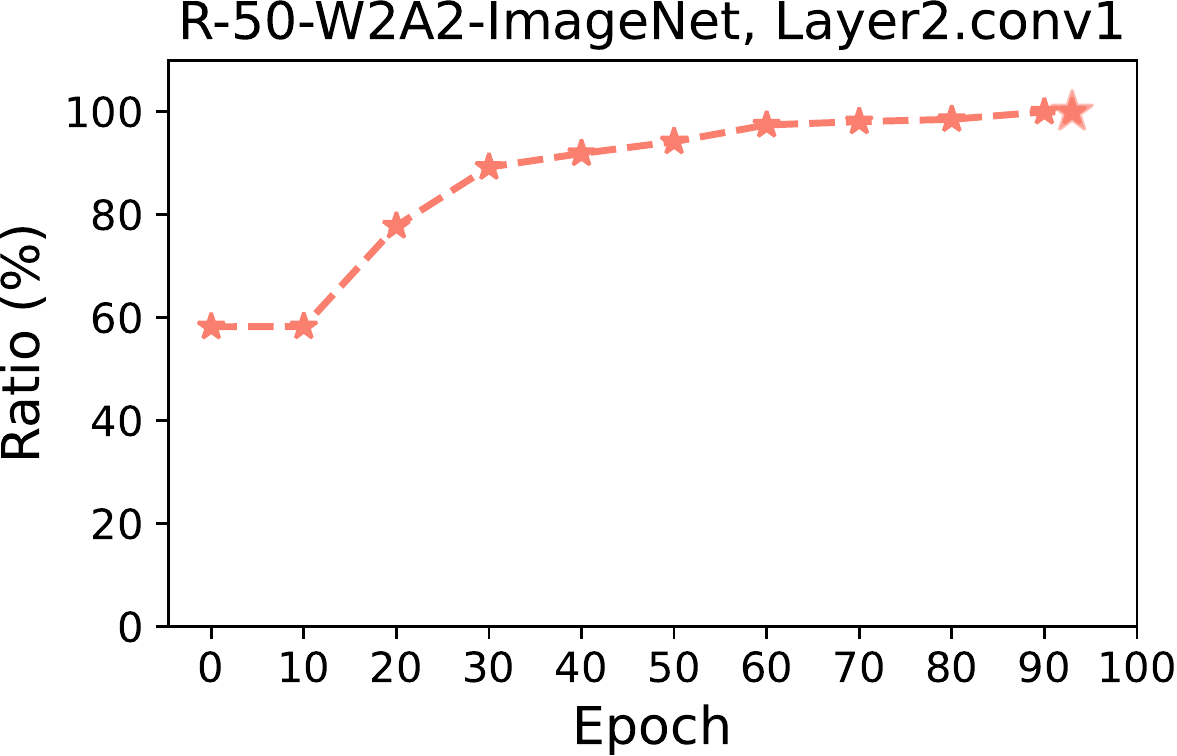}
  \label{insight:g}}
\subfloat[]{
  \includegraphics[width=0.32\linewidth]{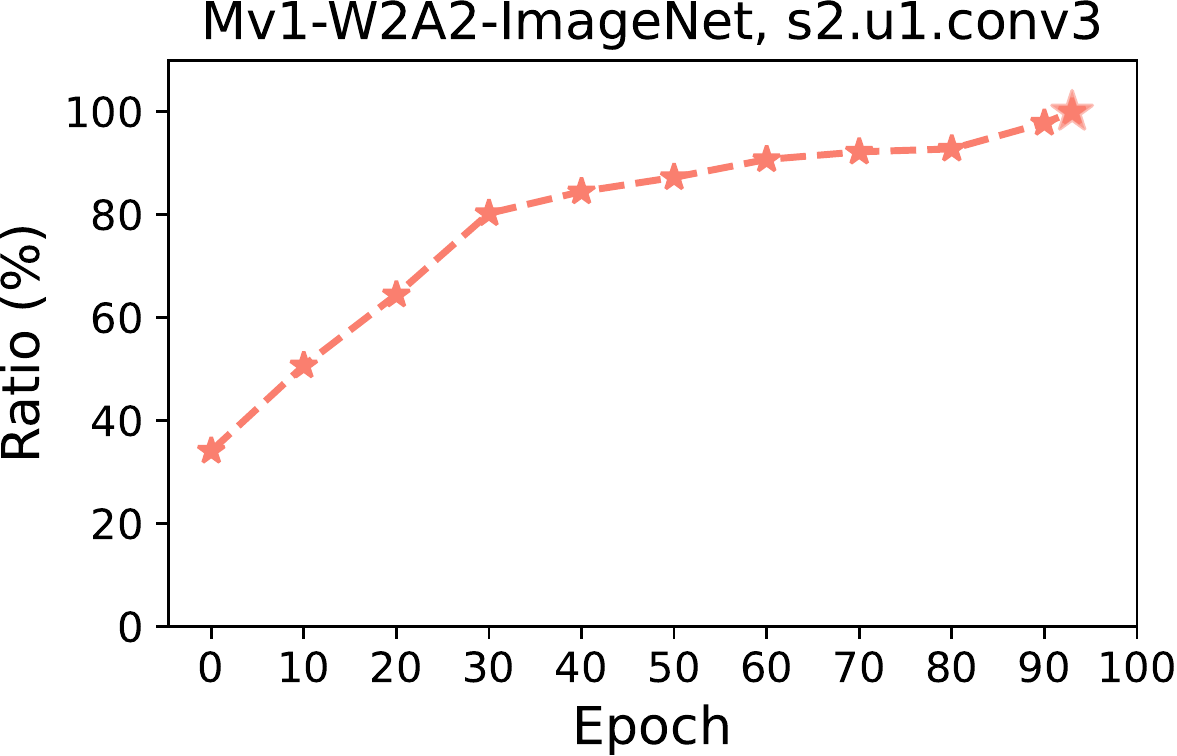}
  \label{insight:h}}
\subfloat[]{
  \includegraphics[width=0.32\linewidth]{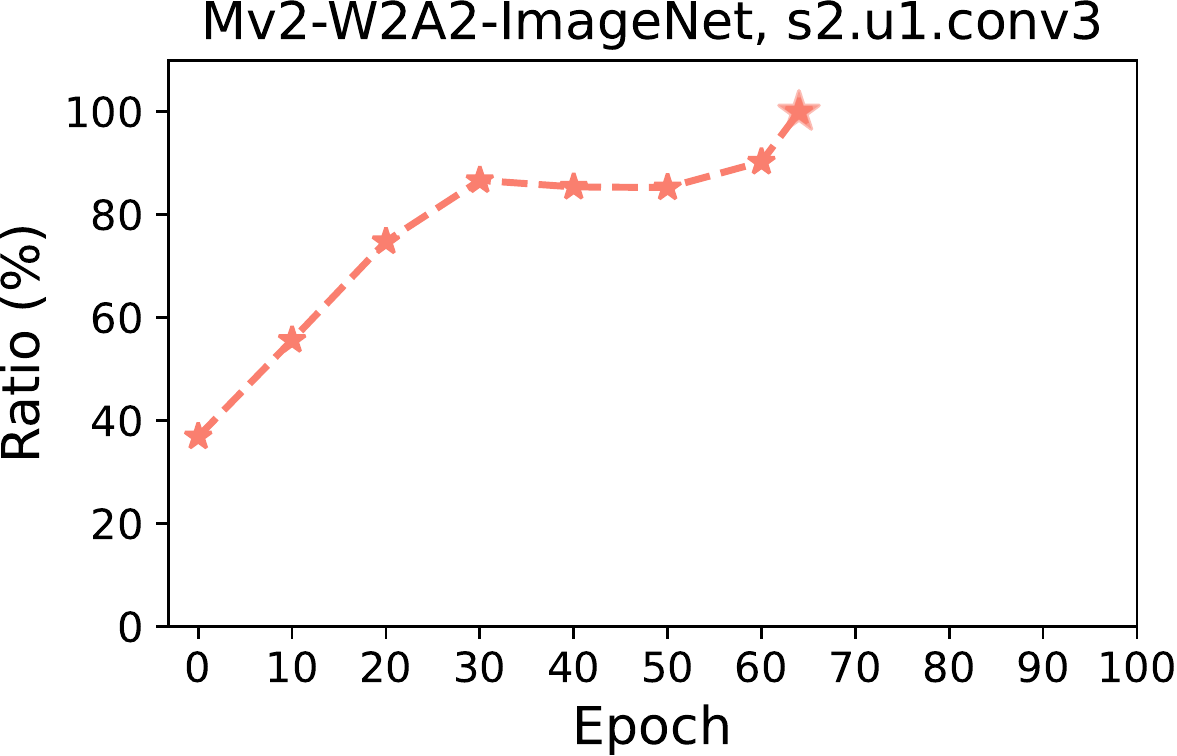}
  \label{insight:i}}
\caption{The ratio of the weights reaching the optimal quantization level \emph{w.r.t.} training epochs. The red pentagon denotes the best model. ``WBAB'' indicates the weights and activations are quantized to B-bit. ``R-20'', ``R-18'', ``R-50'', ``Mv1'', and ``Mv2'' indicate ResNet-20, ResNet-18, ResNet-50, MobileNetV1, and MobileNetV2, respectively.}
\label{insight}
\end{figure*}

In spite of the merits, quantized networks easily suffer from severe performance drops if simply performing the low-bit quantization, since the low-bit data format possesses a very limited representation capacity compared with the full-precision counterpart. 
Thus, the quantized network usually needs to be trained for dozens of epochs to compensate for the performance drop, which is known as quantization-aware training (QAT)~\cite{IntegerOnly}. 
Specifically, in the forward pass, weights and activations are quantized to simulate quantized inference. In the back propagation, the weights are updated as usual. 
All weights of the network are still represented in the full-precision structure, so that the weights can be updated by small gradients.
By continuous updating, the network weights are adjusted to accommodate the quantization effects, which therefore alleviates the performance drop significantly.
In QAT, the common experience is that all quantized weights are updated throughout the entire training period.
On the contrary, in this paper, we challenge the necessity of continuous weight updating: Whether it is necessary to update all weights in the whole training process of QAT?
\IEEEpubidadjcol

To this end, we dive into the training evolution of the quantization level for each weight in QAT. For the first time, we identify an interesting observation that a large portion of network weights quickly reach their optimal quantization levels after a few training epochs. 
As shown in Fig.\,\ref{insight:a}, for ``layer1.conv1'' of the 2-bit ResNet-20~\cite{he2016deep} trained on CIFAR-100~\cite{cifar}, over 50\% weights have already reached their optimal quantization levels even without any training, and the portion rises up to 80\% at epoch 80.
For ``layer1.conv1'' of the 2-bit ResNet-20 on CIFAR-10~\cite{cifar}, near 50\% over have already reached their optimal quantization levels before training and 80\% weights reach their optimal quantization levels at epoch 120 as exhibited in Fig.\,\ref{insight:b}.
Similar observations can be found from Fig.\,\ref{insight:c}-Fig.\,\ref{insight:f} across different bit widths and datasets.
Inspired by the lottery ticket hypothesis in network pruning~\cite{frankle2018lottery}, in this paper, we term these weights that quickly reach the optimal quantization levels as the partly scratch-off lottery ticket since they appear along with training.
We attribute this phenomenon to two characteristics in network quantization.
First, a network is usually quantized from a well-trained full-precision model, which gives good initial values.
Second, a discrete quantization level often covers a range of continuous numerical values.
Thus, many weights rapidly converge to their best quantization levels although their full-precision version still remains far away from the optimal states.

\begin{figure*}[!t]
\centering
\subfloat[]{
\includegraphics[width=0.58\linewidth]{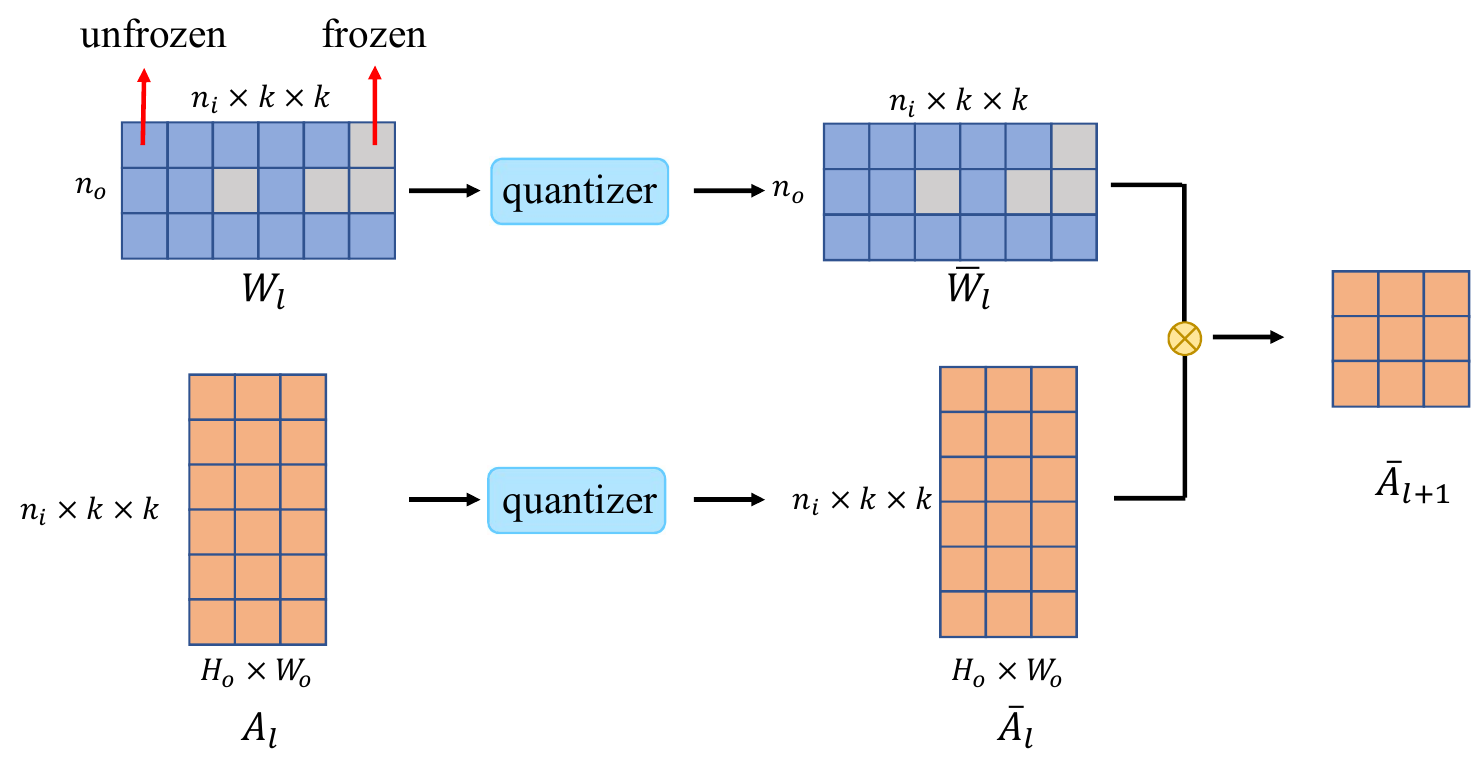}%
\label{illu:forward}}
\vspace{2em}
\subfloat[]{
\includegraphics[width=0.38\linewidth]{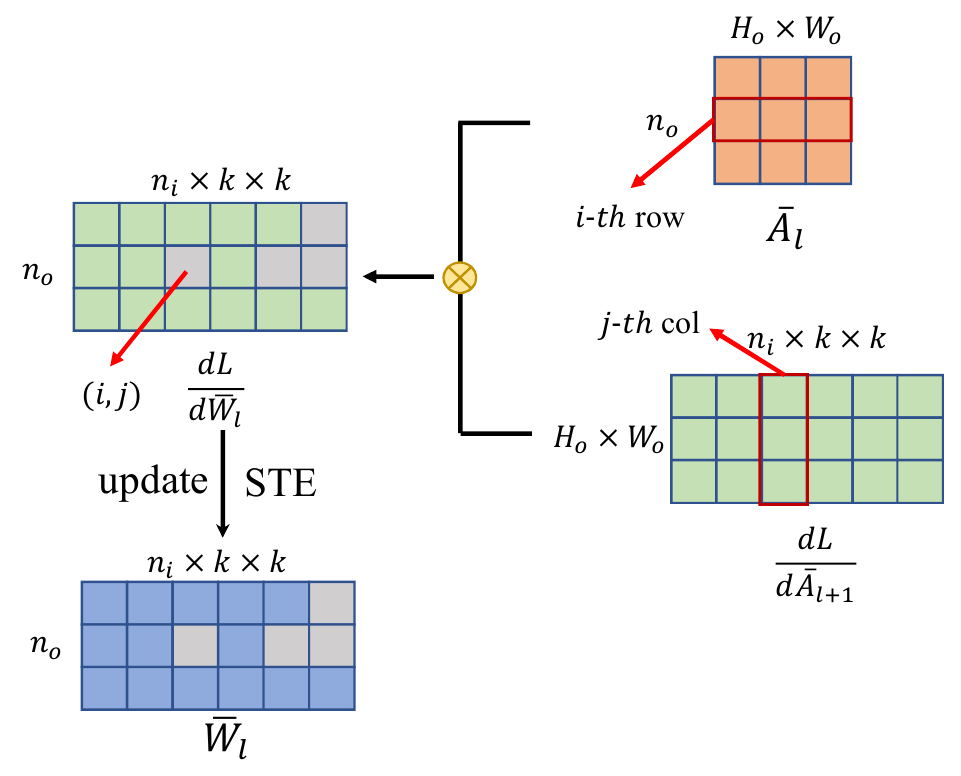}%
\label{illu:backward}}
\caption{An illustration of LTS. For a better illustration, the activations are represented in a unfold manner (\textit{im2col} operation). (a) Frozen and unfrozen weights in the forward pass. (b) Sparsifying weight gradients in the backward pass.}
\label{illustration}
\end{figure*}

Such an observation indicates that a portion of weights can be safely frozen and their gradients do not require calculations throughout the whole training process.
To find the partly scratch-off lottery ticket, we introduce a simple-yet-effective heuristic method, named lottery ticket scratcher (LTS).
Specifically, we first measure the distance between a normalized weight and its corresponding quantization level. Then, the partly scratch-off lottery ticket comprises these weights whose exponential moving average (EMA) distance is lower than a pre-defined threshold.
A weight will be permanently frozen once it falls into the partly scratch-off lottery ticket.
An illustration of our LTS is presented in Fig.\,\ref{illustration}. In the forward pass, the full-precision weights and activations are quantized by the quantizer, the results of which are multiplied to derive the outputs.
In the backward pass, we freeze these weights within the partly scratch-off lottery ticket by setting their gradients to zero, which actually leads to a sparse weight gradient.
This sparse weight gradient indicates a reduction in the training costs, since the related calculations of these gradients can be avoided. 
Furthermore, such a sparsity in weight gradient incurs an easy-implemented structured computational reduction as stated in Sec.\,\ref{sec:imple}.

We also emphasize that LTS is fundamentally different from sparse training methods in four aspects~\cite{bellec2018deep,mostafa2019parameter}.
First, our LTS focuses on a sparse weight gradient, which involves a structured computational reduction (See Sec.\,\ref{sec:imple}).
Second, LTS does not prune any weight while sparse training prunes weights. Third, LTS stops updating frozen weights forever while sparse training usually revives weights, which involves continuous weight gradient calculations~\cite{liu2021sparse}. Lastly, LTS is specialized in quantized networks while sparse training is for full-precision models.

By precisely identifying the partly scratch-off lottery ticket, the proposed LTS not only achieves comparable or even better performance than the compared baseline, but also results in a non-negligible reduction in training costs.
Generally, LTS achieves comparable performance across various quantization, and even improves the performance when quantizing the network in the same low-bit cases for our LTS effectively solves the weight oscillations problem~\cite{pmlr-v162-nagel22a} (Sec.\,\ref{sec:res-cls}).
At the same time, LTS typically obtains an average of 30\%-60\% weight gradient sparsity, which eliminates a total of 15\%-30\% FLOPs of backward propagation.
Taking 2-bit ResNet-18 as an example, compared with the normal QAT, our LTS improves performance by 1.41\% while eliminating 56\% weight updating and 28\% FLOPs of the backward pass.

\section{Related Work}

\subsection{Quantization-Aware Training}
Network quantization has long been the research focus of the model compression community for its superiority in simultaneously reducing the model size and computational costs.
However, the quantized network often suffers from performance degradation. A variety of techniques are explored to overcome this issue.
Most existing QAT methods focus on designing quantizers that are more suitable for the network weights or activations~\cite{LSQ,APoT,jung2019learning,liu2022nonuniform,zhang2021differentiable}. Other methods instead pay attention to special training strategies for quantized networks~\cite{zhou2016incremental,zhuang2020training,lee2021network}, approximating gradients of the quantization functions~\cite{gong2019differentiable,kim2021distance}, training regularization~\cite{lee2021cluster,han2021improving} or mix-precision quantization~\cite{dong2019hawq,yang2021fracbits,chen2021towards}. 
Although these techniques alleviate performance degradation, they stiffly rely on the training overhead.

\subsection{Lottery Ticket Hypothesis}
The lottery ticket hypothesis is first proposed in~\cite{frankle2018lottery}, which indicates that a randomly initialized network contains sub-networks (\emph{a.k.a.}, ``winning tickets'') that can achieve the test accuracy of the original network when trained independently.
Following this conjecture, many empirical studies~\cite{zhou2019deconstructing,ramanujan2020s, zhang2021lottery,wang2020picking,pensia2020optimal} have been devoted to finding the lottery ticket. 
In contrast, the partly scratch-off lottery ticket in this paper is different. We firstly reveal the existence of a sub-network that does not require updating in QAT, while the lottery ticket targets weight pruning weight and full-precision networks.

\subsection{Training Acceleration}
Many efforts have been made to accelerate the training DNNs.
One common way is to exploit the highly efficient distributed training~\cite{goyal2017accurate,jia2018highly,you2018imagenet,akiba2017extremely}. 
However, the distributed training is towards modern single- and multi-GPU servers, clusters, and even supercomputers, where the acceleration comes at expensive training costs and energy consumption.
In contrast, our LTS aims to reduce training costs.

Gradient pruning targets pruning gradients in the computationally-intensive backpropagation~\cite{sun2017meprop,goli2020resprop}. For example, meProp~\cite{sun2017meprop} preserves the top-$k$ gradients in the matrix multiplication output.
In contrast to gradient pruning which continuously updates weights in the training process, our LTS stops updating these frozen weights forever.

Sparse training imposes sparsity constraints during network training~\cite{mostafa2019parameter,evci2020rigging,liu2021sparse,raihan2020sparse,bellec2018deep,evci2021gradmax,liu2021we}. By continuously removing and reviving network weights in the training, related computations in forward and backward propagations can be reduced. Also, these methods are often designed for full-precision networks.
Differently, our LTS does not remove or revive any weight and is particularly designed for network quantization.

Another line of work represents gradients in a low-bit data format to reduce gradient computing costs~\cite{banner2018scalable,zhu2020towards}. In compliance with extensive analyses on gradient distribution, most current methods focus on designing a suitable quantizer~\cite{zhao2021distribution}. For instance, Lee~\emph{et al.}~\cite{lee2021toward} implemented a low-bit representation of gradients in a searching-based style. \cite{chmiel2020neural} reasoned a near-lognormal distribution of gradients and obtained the best clipping value analytically.
Our LTS is orthogonal to these methods since we focus on quantizing the network in the forward pass. These methods can be integrated into our LTS to further speed up the network training.

\section{Method}

\subsection{Preliminary}
\label{sec:Preliminary}

%

%
A quantizer is adopted in QAT to quantize the full-precision weights and activations to simulate the quantization inference. 
Following the settings of~\cite{lee2021network,hu2022palquant}, we use the uniform quantizer with trainable clipping parameters to implement network quantization. Concretely, given full-precision data $\bm{x}$ (weights or activations), it is first normalized to a range $[0, 1]$ with two trainable clipping parameters $l, u$ by the following equation:
\begin{equation}
\bm{x}_n = clip(\frac{\bm{x}-l}{u-l}, 0, 1),
\label{quantizer1}
\end{equation}
where $clip(\bm{x_n}, l, u) = min\big(max(\bm{x_n}, l), u\big)$, $ \lfloor \cdot \rceil$, $l, u$ denotes lower bound and upper bound, respectively. Then, the normalized data $\bm{x_n}$ is quantized by the following quantizer:
\begin{equation}
\bm{q} = \lfloor (2^b - 1) \times \bm{x}_n)\rceil,
\label{quantizer2}
\end{equation}
where $\lfloor \cdot \rceil$ denotes the round function that rounds its input to the nearest integer. The corresponding de-quantized value $\bar{\bm{x}}$ is calculated according to its data type. For weights, the de-quantized value is calculated as:
\begin{equation}
    \bar{\bm{x}}  =  2 \times \big (\frac{\bm{q}}{2^b-1} - 0.5 \big),  \bm{x} \in weights. 
\end{equation}

For activations, the de-quantized value is calculated as:
\begin{equation}
    \bar{\bm{x}}  =  \frac{\bm{q}}{2^b-1},  \bm{x} \in activations. 
\end{equation}

For activations and weights, we both use layer-wise quantizer.
Following~\cite{hu2022palquant}, for weight, the lower bound $l$ and upper bound $u$ are initialized to negative three times the standard deviation and negative three times the standard deviation of the full-precision weight, respectively. For activations, the lower bound $l$ and upper bound $u$ are initialized to minimum and maximum values of the full-precision activations, respectively.

\subsection{Partly Scratch-off Lottery Ticket}

The limited capacity of low-bit representation requires accomplishing network quantization in a quantization-aware training (QAT) manner so as to mitigate the performance gap between the quantized network and its full-precision counterpart. As a common experience, the weights of the quantized network are updated for the entire training process in QAT.
On the contrary, in this paper, we challenge the necessity of updating weight throughout the whole training process of QAT, which is barely investigated in the literature.

To this end, we dive into the training evolution of quantization level for each weight and unearth an interesting phenomenon that many network weights converge to their optimal quantization levels in the early training process. 
%
%
%
Concretely, we calculate the proportion of quantized weights that reach the optimal quantization level along with the network training.
Fig.\,\ref{insight} provides the results over various datasets and bit-widths.
Taking Fig.\,\ref{insight:a} as an instance, for ``layer1.conv1'' of 2-bit ResNet-20 on CIFAR-100, the optimal quantized network is obtained at around epoch 320.
However, over 65\% of weights reach their optimal quantization levels without any training. And it increases to 80\% at epoch 80.
A similar observation can be found at other bit-widths. For example, in Fig.\,\ref{insight:d}, 20\% weights already stay in the optimal quantization level before training a 4-bit ResNet-20 on CIFAR-100 and it becomes 40\% at epoch 80.
Networks on larger datasets such as ImageNet~\cite{russakovsky2015imagenet} also present a similar observation as that in CIFAR-10/100. For instance, for 2-bit ResNet-18, over 50\% weights stay in the optimal quantization level before training and it becomes 60\% at epoch 10.

The above phenomenon indicates that the half-trained quantized network contains a quantized subnet consisting of these well-optimized quantized weights. Alike to the lottery ticket in network pruning~\cite{frankle2018lottery}, we term these weights as the partly scratch-off lottery ticket, which means they do not require to be trained from the beginning to end. 
When diving into a deeper analysis, we attribute the existence of the partly scratch-off lottery ticket to two possible characteristics in network quantization.
First, weights of the quantized network are often initialized from a pre-trained full-precision model. It can be expected that many weights start from an optimal or sub-optimal state. 
As a result, several training epochs lead to the convergence of many weights in their optimal quantization level.
Second, compared with the full-precision data format, the quantization levels are discrete. 
%
%
Numerous continuous values in an interval are mapped to the same discrete state. Despite that full-precision weights vary across different training epochs, they remain at the same quantization level.
In particular, for the lower bit-width case, more partly scratch-off lottery tickets can be observed since each interval becomes much larger. Fig.\,\ref{insight} demonstrates this potential: the 2-bit ResNet-20 manifests a higher ratio of the optimal quantization level than its 4-bit version in the early training process.

Based on this phenomenon, we realize the unnecessary expense of updating these tickets during most part of the training period since they already fall into their optimal quantization levels. 
Therefore, if we can well locate these tickets, it can be expected that the performance degradation of the quantized network is negligible.
Moreover, as an intuitive byproduct, we can reduce the training costs by creating a weight gradient sparsity due to the gradient calculations of these tickets can be eliminated.

\subsection{Lottery Ticket Scratcher}
In this subsection, we introduce lottery ticket scratcher (LTS), a heuristic method, to discover the partly scratch-off lottery. By gradually pulling out weights from network training, the proposed LTS creates a sparse weight gradient, leading to a shrinkage of training costs.

Specifically, our LTS employs a simple-yet-effective heuristic rule to find the partly scratch-off lottery ticket. 
At first, we train the quantized network for $E_{wm}$ epochs as the warmup stage.
Then, the partly scratch-off lottery comprises weights whose EMA distance is lower than a pre-defined threshold $t$. Herein, the EMA distance is the exponential moving average distance between the normalized weight $\bm{w_n}$ and its corresponding quantization level. At the $i$-th training iteration, the EMA distance of the given $\bm{w_n}$ is defined as: 
\begin{equation}
D^i_{\bm{w_n}} =  m D^{i-1}_{\bm{w_n}} + (1 - m) D_{\bm{w_n}},
\label{equ:emadistance}
\end{equation}
where $m$ is the momentum of the EMA function, and $D_{\bm{w_n}}$ is the distance between $\bm{w_n}$ and its corresponding quantization level. Note that $D^i_{\bm{w_n}}$ will be reset to the quantization interval $\Delta^B = \frac{2}{2^B}$ once the quantization level of $\bm{w_n}$ is changed. 
We then compare $D^i_{\bm{w_n}}$ with a threshold $t$ to determine whether $\bm{w_n}$ should be frozen. If $D^i_{\bm{w_n}}$ is lower than the threshold $t$, $\bm{w_n}$ will be frozen, meaning that zeroing out the gradient of $\bm{w_n}$. Otherwise, $\bm{w_n}$ is continuously updated.
The threshold $t$ is computed by obtained $\Delta^B$ with a rate $p$:
\begin{equation}
t =  \Delta^B \cdot p.
\label{equ:threshold}
\end{equation}

The constant quantization interval $\Delta^B$ is determined by the bit-width $B$. We adjust the rate $p$ to control the ease of frozen weights.
The selection of $p$ provides a trade-off between performance and training costs reduction.
For example, a small $p$ returns a small threshold $t$, which means a weight has to be very close to its quantization level if it is frozen.
As a result, the identification of the partly scratch-off lottery is more accurate that is able to retain the performance. However, it is more difficult to freeze weights and more weights are prone to be updated. In this case, the weight gradient sparsity is limited and the reduced training costs are also limited.
In contrast, a large $p$ indicates a large threshold $t$.
Subsequently, weights are more likely to be misidentified as the partly scratch-off lottery since the condition is loose. This misidentification will lead to performance degradation.
However, it would be easier to freeze weights and achieve a higher weight gradient sparsity, so that the training costs can be reduced significantly.

To adjust the value of $p$, we design three strategies including fixing, linear-growth, and sine-growth.
The fixing strategy indicates the rate $p$ is set to a constant $c$ that ranges from 0 to 1.
In linear-growth strategy and sine-growth strategy, the rate $p$ gradually increases in a linear manner and a sine manner when the warmup stage finishes, respectively. Supposing the total training epoch is $E$, the current iteration is $i$, and an epoch has $T$ iterations. The linear-growth strategy is defined as:
\begin{equation}
p = \frac{i-T \cdot E_{wm}}{T \cdot E - T \cdot E_{wm}} \cdot \mathbb{I}{(i > T \cdot E_{wm})},
\label{equ:linear-growth}
\end{equation}
where $\mathbb{I}()$ is the indicator function. The sine-growth is defined as:
\begin{equation}
p =   \sin \big ( \frac{i-T \times E_{wm}}{T \cdot E - T \cdot E_{wm}} \cdot \frac{\pi}{2} \big ) \cdot \mathbb{I}{(i > T \cdot E_{wm})}.
\label{equ:sine-growth}
\end{equation}

Both linear-growth strategy and sine-growth strategy increase the rate $p$ from 0 to 1.
We employ the linear-growth strategy in our experiments for it achieves a good balance between sparsity and accuracy (Sec.\,\ref{sec:res-cls}).

\subsection{Efficient Implementation}
\label{sec:imple}
In this subsection, we demonstrate that the sparse weight gradients actually lead to a structured computational reduction, which is easy to implement.
The illustration of the backward pass is shown in Fig.\,\ref{illu:backward}.
As can be seen, the gradient of weight in position $(i, j)$ is computed by multiplying the $i$-th row of the $\bar{A}_l$ and the $j$-th column of the $\frac{\partial L}{\partial \bar{A}_{l+1}}$.
The computational process is structured since it only involves the multiplication of two vectors.
Due to we already know the position of the frozen weight before weight gradient calculations, we can omit the vector multiplications according to the set consisting of positions of frozen weights.
We implement it with the CUDA toolkit and the result shows that 50\% average weight gradient sparsity gives us roughly 20\% time cost reduction in the backward propagation.

\section{Experimentation}

\subsection{Experimental Settings}

\subsubsection{Implementation}

We quantize ResNet-20~\cite{he2016deep} for CIFAR-100/10~\cite{cifar}, ResNet-18, ResNet-50~\cite{he2016deep}, MobileNetV1~\cite{howard2017mobilenets}, and MobileNetV2~\cite{sandler2018mobilenetv2} for ImageNet~\cite{russakovsky2015imagenet}. All code are implemented with Pytorch~\cite{paszke2019pytorch}. 
For all experiments we quantized all layers of networks into 2-, 3-, and 4-bit.

All networks are trained with cross-entropy loss.
The gradient of the round function is set to 1 by using the straight-through estimator (STE)~\cite{courbariaux2016binarized}.
The SGD optimizer is adopted with a momentum of 0.9.
For CIFAR-100/10, the initial learning rate, batch size, weight decay, and total training epochs are set to 0.1, 256, 0.0001, and 400, respectively. 
For ImageNet, the batch size, weight decay, and total training epochs are set to 256, 0.0001, and 100, respectively.
The initial learning rate is set to 0.1 for ResNet-18/50 and 0.01 for MobileNetV1 and MobileNetV2, respectively.
The learning rate is decayed by a factor of 0.1 for every 100 epochs on CIFAR-100/10, and every 30 epochs on ImageNet.
The pre-trained MobileNetV1 is obtained from \textit{pytorchcv} and other models are downloaded from the \textit{torchvision}.
For all experiments, we employ the linear-growth strategy for it achieves the best trade-off between sparsity and accuracy.
For the hyper-parameters, we use grid search to search for the best configuration for each dataset. We set $m$ as 0.99 for CIFAR-100/10, and as 0.9999 for ImageNet. For $E_{wm}$, we set it as 80 for CIFAR100/10, and as 30 for ImageNet by using the grid search. 
Despite they might not be optimal for all networks, we find these values already have provided satisfactory performance.

\begin{figure*}[!t]
\centering
  \subfloat[xx]{
  \includegraphics[width=0.45\linewidth]{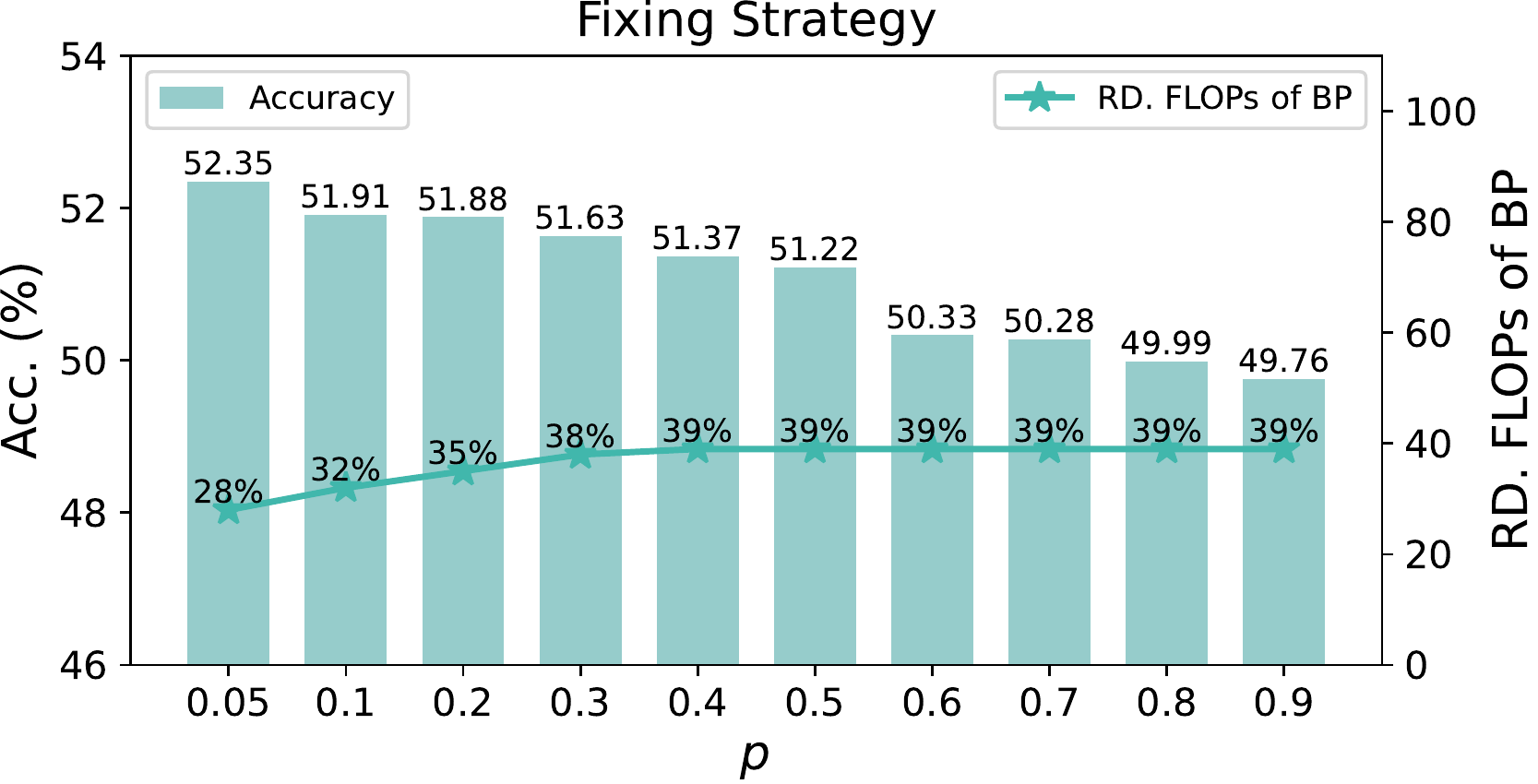}
  \label{ablation:a}}
  \subfloat[xx]{
  \includegraphics[width=0.45\linewidth]{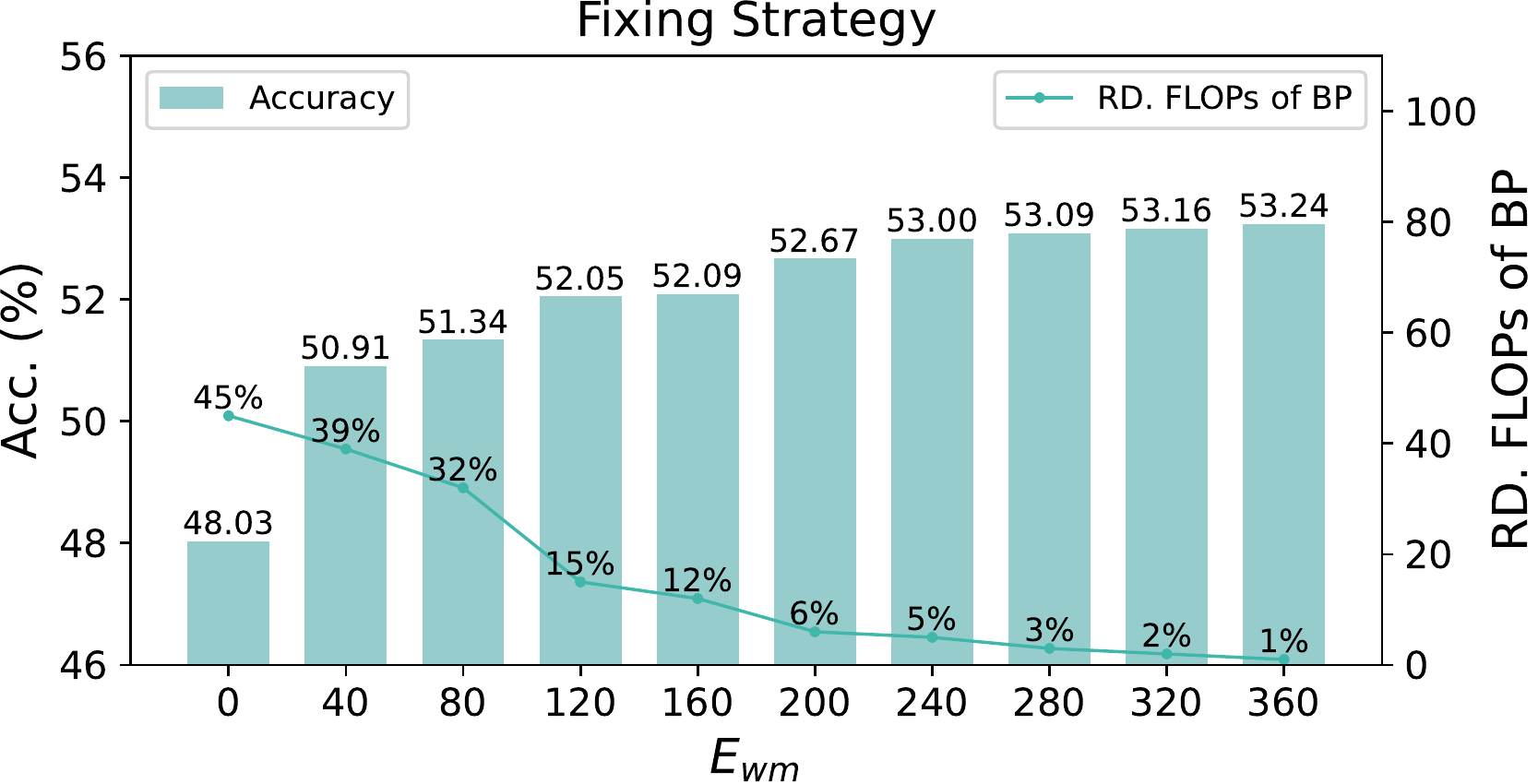}
  \label{ablation:b}}
    \\
  \subfloat[xx]{
  \includegraphics[width=0.45\linewidth]{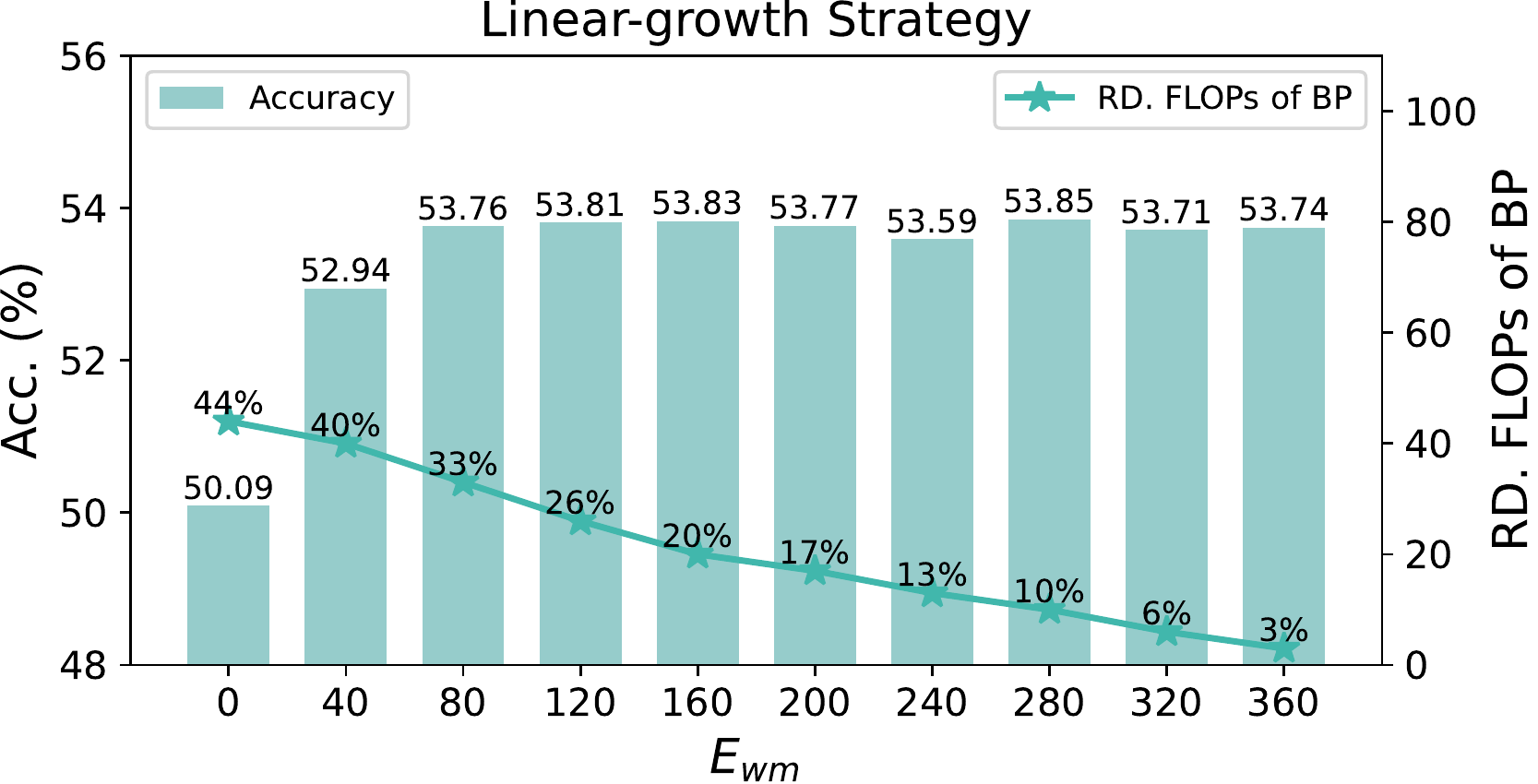}
  \label{ablation:c}}
  \subfloat[xx]{
  \includegraphics[width=0.45\linewidth]{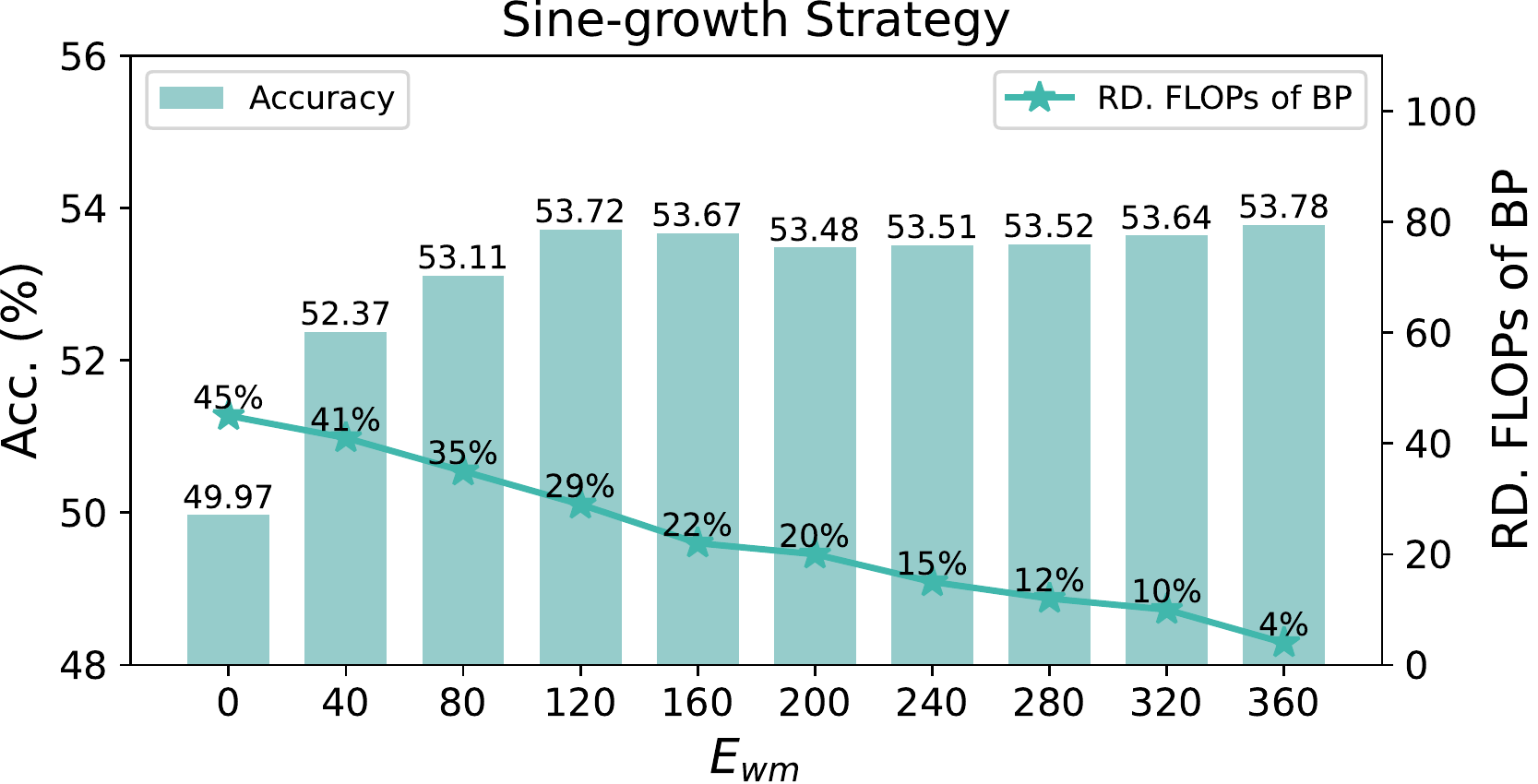}
  \label{ablation:d}}
\caption{Influence of the hyper-parameters on the top-1 accuracy of 2-bit ResNet-20 on CIFAR-100. (a) and (b) provide the ablation studies of the fixing strategy. (c) and (d) provide the ablation studies of the linear-growth strategy and sine-growth strategy, respectively. ``RD. FLOPs of BP'' indicates the reduction in FLOPs of backward propagation.}
\label{ablation}
\end{figure*}

\begin{table*}[ht]
\centering
\caption{Results of ResNet-20 on CIFAR-100/10. ``Avg. of WGS'' indicates the average weight gradient sparsity. ``RD. FLOPs of BP'' indicates the reduction in FLOPs of backward propagation.}
\label{tab:cifar-100/10}
\begin{tabular}{ccccccc}
\toprule[1.25pt]
Datasets                  & Networks                   & W/A                  & Mode     & Accuracy. (\%) & Avg. of WGS & RD. FLOPs of BP   \\ \hline \hline
\multirow{6}{*}{CIFAR-100} & \multirow{6}{*}{\begin{tabular}[c]{@{}c@{}}ResNet-20\\ (FP: 65.24\%)\end{tabular}} & \multirow{2}{*}{2/2} & Baseline & 52.95     &  0\%                      &    0\%         \\
                          &                            &                      & \textbf{LTS} (Ours)    &   \textbf{53.76}    &   66\%               &       33\%$\downarrow$      \\ \cline{4-7} 
                          &                            & \multirow{2}{*}{3/3} & Baseline &   62.91   &   0\%                  &        0\%     \\
                          &                            &                      & \textbf{LTS} (Ours)    &    \textbf{63.67}   &  68\%             &    34\%$\downarrow$         \\ \cline{4-7} 
                          &                            & \multirow{2}{*}{4/4} & Baseline &  65.44  &     0\%                  &    0\%         \\
                          &                            &                      & \textbf{LTS} (Ours)    & \textbf{66.43}    &   70\%                   &   35\%$\downarrow$      \\ \hline
\multirow{6}{*}{CIFAR-10}  & \multirow{6}{*}{\begin{tabular}[c]{@{}c@{}}ResNet-20\\ (FP: 91.46\%)\end{tabular}} & \multirow{2}{*}{2/2} & Baseline &  84.80   &   0\%                    &        0\%     \\
                          &                            &                      & \textbf{LTS} (Ours)    &   \textbf{85.33} &  69\%                  &    34.5\%$\downarrow$         \\ \cline{4-7} 
                          &                            & \multirow{2}{*}{3/3} & Baseline & 89.99    &     0\%                   &    0\%         \\
                          &                            &                      & \textbf{LTS} (Ours)    &   \textbf{90.58}   &      69\%         &       34.5\%$\downarrow$      \\ \cline{4-7} 
                          &                            & \multirow{2}{*}{4/4} & Baseline & \textbf{91.71}  &     0\%              &    0\%         \\
                          &                            &                      & \textbf{LTS} (Ours)    &    91.70    &        70\%             &      35\%$\downarrow$       \\ \bottomrule[0.75pt]
\end{tabular}
\end{table*}

\subsubsection{Metrics}

We report the top-1 accuracy as the metric.
To measure the reduced computational training costs, we report the average weight gradient sparsity, which is the sum of the weight gradient sparsity per iteration divided by the total number of iterations.
The weight gradient sparsity per iteration is the ratio of the number of frozen weights to the number of all weights at the current iteration. 
We also employ the overall reduction in FLOPs of backward propagation to measure training cost reduction.
This metric is equal to half of the average weight gradient sparsity since LTS only eliminates the computation of weight gradient.

\subsection{Ablation Study }
\label{app:ablation}

\begin{table}[ht]
\centering
\caption{Ablation study of the EMA parameter $m$.}
\label{tab:ema}
\begin{tabular}{l|ccc}
\toprule[1.25pt]
$m$       & Accuracy. (\%)     & Avg. of WGS & RD. FLOPs of BP  \\ \hline \hline
0 & 52.43\% &  \textbf{73\%}  &  \textbf{36.5\%$\downarrow$} \\ \hline
0.9 & 53.06\% & 73\%  & 36\%$\downarrow$ \\ \hline
0.99 & 53.76\% & 66\% &  33\%$\downarrow$ \\ \hline
0.999 & 53.81\% & 60\% &  30\%$\downarrow$ \\ \hline
0.9999 & 53.84\% & 51\% & 25.5\%$\downarrow$ \\ \bottomrule[0.75pt]
\end{tabular}
\end{table}
 
In this section, we first conduct ablation studies of the proposed three strategies and hyper-parameters of our LTS. All experiments are conducted by quantizing ResNet-20 to 2-bit on CIFAR-100. The top-1 accuracy and the reduction in FLOPs of backward propagation are reported.

\textbf{EMA Parameter.} 
The ablation study of EMA parameter $m$ is shown in Tab.\,\ref{tab:ema}. When $m=0$, it is equal to EMA not used.
It can be seen that the highest average weight gradient sparsity is achieved when the $m=0$, which is 73\%. However, compared with $m=0.99$, the 7\% improvement on average weight gradient sparsity comes at the cost of 1.33\% accuracy degradation.
As the $m$ increases, the accuracy improvement is slight while the average weight gradient sparsity decreases quickly. Specifically, compared to the results of $m=0.99$, results of $m=0.999$ only lead to 0.05\% accuracy gain but the sparsity decreases by 6\%.

\textbf{Fixing Strategy.} 
The fixing strategy has two hyper-parameters including $p$ and $E_{wm}$. Fig.\,\ref{ablation:a} and Fig.\,\ref{ablation:b} respectively provide the top-1 accuracy of 2-bit ResNet-20 on CIFAR100 \emph{w.r.t} $p$ and $E_{wm}$. 
From Fig.\,\ref{ablation:a}, it can be observed that the reduction in FLOPs of backward propagation increases along with the increase of $p$, and the best accuracy is obtained when $p=0.05$.
As presented in Fig.\,\ref{ablation:b}, for fixing strategy, it is hard to find a good trade-off between accuracy and backward propagation FLOPs reduction since the accuracy improvements are accompanied by a low FLOPs reduction. Thus, compared with the other two strategies, the fixing strategy does not exhibit any advantage.

\textbf{Linear-growth Strategy.} 
As illustrated in Fig.\,\ref{ablation:c}, as the warmup epochs $E_{wm}$ increases, the backward propagation FLOPs reduction consistently decreases. The performance reaches 53.76\% and the FLOPs reduction is 33\% at $E_{wm}=80$.
When $E_{wm}>80$, the accuracy has not been greatly improved or even decreased.

\textbf{Sine-growth Strategy.}
From Fig.\,\ref{ablation:d}, we can find that the accuracy is 53.72\% when $E_{wm}=120$. After that, the performance improvement is marginal or negative despite the backward propagation FLOPs reduction decreasing rapidly.

In summary, comparing these three strategies, we can observe that the linear-growth strategy and the sine-growth strategy obtain a better performance than the fixing strategy. Moreover, the linear-growth strategy enjoys higher FLOPs reduction than the sine-growth when they have comparable accuracy. 
Thus, we consider it achieves the best trade-off between performance and the backward propagation FLOPs reduction and thus employ this strategy in all following experiments.

\begin{table*}[ht]
\centering
\caption{Results of ResNet-18, ResNet-50, MobileNetV1, and MobileNetV2 on ImageNet.}
\label{tab:ImageNet}
\begin{tabular}{ccccccc}
\toprule[1.25pt]
Datasets                   & Networks                    & W/A                  & Mode     & Accuracy. (\%) & Avg. of WGS & RD. FLOPs of BP   \\ \hline \hline
\multirow{24}{*}{ImageNet} & \multirow{6}{*}{\begin{tabular}[c]{@{}c@{}}ResNet-18\\ (FP: 69.64\%)\end{tabular}} & \multirow{2}{*}{2/2} & Baseline & 55.04    &   0\%       &    0\%      \\
                           &                             &                      & \textbf{LTS} (Ours)    &  \textbf{55.32}   &    49\%     &   24.5\%$\downarrow$    \\ \cline{4-7} 
                           &                             & \multirow{2}{*}{3/3} & Baseline & 66.69  &    0\%        &    0\%       \\
                           &                             &                      & \textbf{LTS} (Ours)    & 66.28  &   51\%      &  25.5\%$\downarrow$   \\ \cline{4-7} 
                           &                             & \multirow{2}{*}{4/4} & Baseline &  68.80   & 0\%        &   0\%  \\
                           &                             &                      & \textbf{LTS} (Ours)    & 68.37  & 47\%       &  23.5\%$\downarrow$   \\ \cline{2-7} 
                           & \multirow{6}{*}{\begin{tabular}[c]{@{}c@{}}ResNet-50\\ (FP: 76.15\%)\end{tabular}} & \multirow{2}{*}{2/2} & Baseline & 65.21    &  0\%       &    0\%      \\
                           &                             &                      & \textbf{LTS} (Ours)    & \textbf{65.95}  &   52\%      &   26\%$\downarrow$  \\ \cline{4-7} 
                           &                             & \multirow{2}{*}{3/3} & Baseline &  72.43   & 0\%       &    0\%      \\
                           &                             &                      & \textbf{LTS} (Ours)    & \textbf{72.86}  &   52\%     &   26\%$\downarrow$  \\ \cline{4-7} 
                           &                             & \multirow{2}{*}{4/4} & Baseline & 73.74    & 0\%        &    0\%      \\
                           &                             &                      & \textbf{LTS} (Ours)    &  \textbf{74.19}  &  53\%     &   26.5\%$\downarrow$    \\ \cline{2-7}
                           & \multirow{6}{*}{\begin{tabular}[c]{@{}c@{}}MobileNetV1\\ (FP: 73.33\%)\end{tabular}} & \multirow{2}{*}{2/2} & Baseline &  45.87 &    0\%    &         0\%    \\
                           &                             &                      & \textbf{LTS} (Ours)    &  \textbf{49.42} &    50\%   &      25\%$\downarrow$       \\ \cline{4-7} 
                           &                             & \multirow{2}{*}{3/3} & Baseline & 65.01  &      0\%      &     0\%        \\
                           &                             &                      & \textbf{LTS} (Ours)    & \textbf{66.60} &     51\%       &   25.5\%$\downarrow$    \\ \cline{4-7} 
                           &                             & \multirow{2}{*}{4/4} & Baseline & 70.00 &    0\%     &      0\%       \\
                           &                             &                      & \textbf{LTS} (Ours)    & \textbf{70.50} &    52\%      &     26\%$\downarrow$        \\ \cline{2-7} 
                           & \multirow{6}{*}{\begin{tabular}[c]{@{}c@{}}MobileNetV2\\ (FP: 71.83\%)\end{tabular}} & \multirow{2}{*}{2/2} & Baseline &  40.56 &    0\%    &         0\%    \\
                           &                             &                      & \textbf{LTS} (Ours)    &  \textbf{45.61} &    46\%   &      23\%$\downarrow$       \\ \cline{4-7} 
                           &                             & \multirow{2}{*}{3/3} & Baseline & 63.32 &      0\%      &     0\%        \\
                           &                             &                      & \textbf{LTS} (Ours)    & \textbf{64.14}&     50\%       &   25\%$\downarrow$    \\ \cline{4-7} 
                           &                             & \multirow{2}{*}{4/4} & Baseline & 68.51 &    0\%     &      0\%       \\
                           &                             &                      & \textbf{LTS} (Ours)    & \textbf{69.11} &    52\%      &     26\%$\downarrow$        \\ \bottomrule[0.75pt]
\end{tabular}
\end{table*}

\subsection{Quantitative Comparison}
\label{sec:res-cls}

\subsubsection{CIFAR-100/10}

In this subsection, we first evaluate the performance of the proposed LTS CIFAR-100/10 datasets by comparing them against the baseline. 
The results are presented in Table \ref{tab:cifar-100/10}. Our proposed LTS achieves significant improvements in terms of average weight gradient sparsity and backward propagation FLOPs reduction across all bit-widths, while maintaining consistently high accuracy compared with the baseline.

On CIFAR-100, our proposed LTS achieves an average weight gradient sparsity of 66\%, 68\%, and 70\% for 2-, 3-, and 4-bit, respectively. This corresponds to a reduction in backward propagation FLOPs of 33\%, 34\%, and 35\%. Notably, the proposed LTS improves the performance by 0.81\%, 0.76\%, and 0.99\% for 2-, 3-, and 4-bit, respectively, compared to the baseline.

Similarly, on CIFAR-10, our proposed LTS yields better or comparable performance compared to the baseline across all bit-widths, while achieving high average weight gradient sparsity and backward propagation FLOPs reduction. Specifically, for 2 and 3-bit, the proposed LTS achieves performance gains of 0.53\% and 0.59\%, respectively, and only incurs a minor performance drop of 0.01\% for 4-bit. For 2-, 3-, and 4-bit, LTS respectively provides 69\%, 69\%, and 70\% weight gradient sparsity, which corresponds to a reduction in backward propagation FLOPs of 34.5\%, 34.5\%, and 35\%.

Overall, experimental results demonstrate the superiority of the proposed LTS over the baseline in terms of both accuracy and efficiency on both CIFAR-100 and CIFAR-10 datasets. The significant improvements in average weight gradient sparsity and backward propagation FLOPs reduction across all bit-widths suggest that the proposed LTS is a promising method for improving the efficiency of quantization-aware training.

\begin{figure*}[!t]
\centering
    \subfloat[]{
    \includegraphics[width=0.32\linewidth]{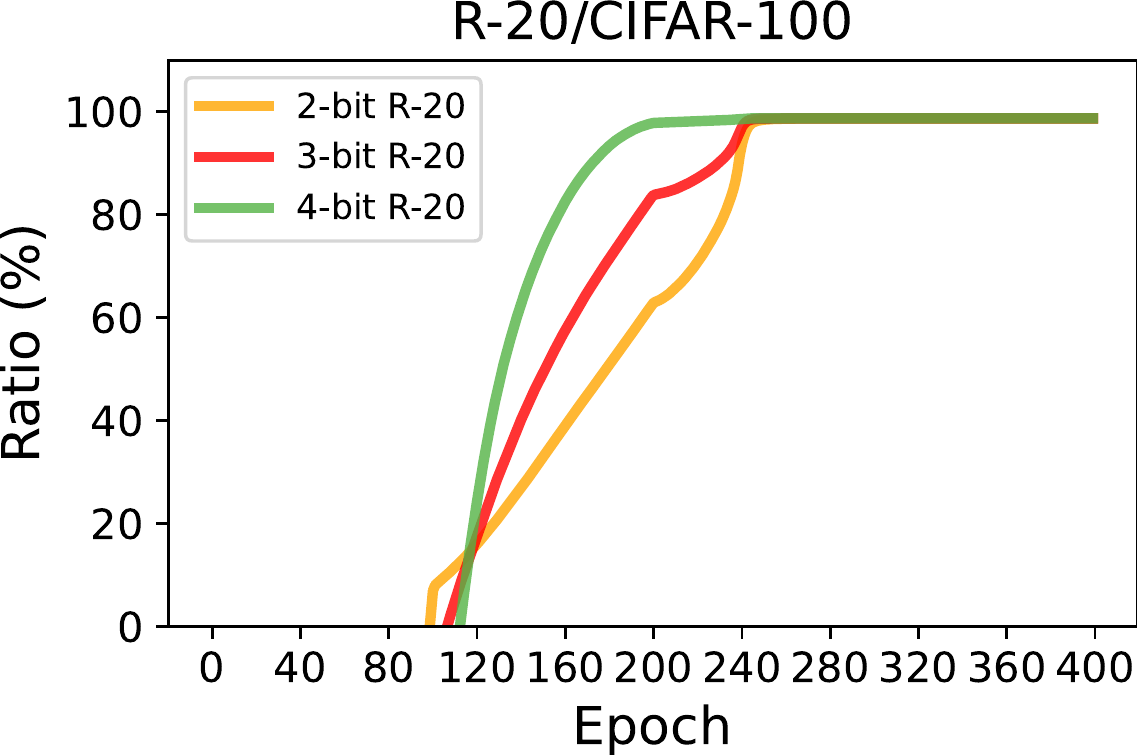}%
    \label{ill-sparsity:a}}
    \subfloat[]{
    \includegraphics[width=0.32\linewidth]{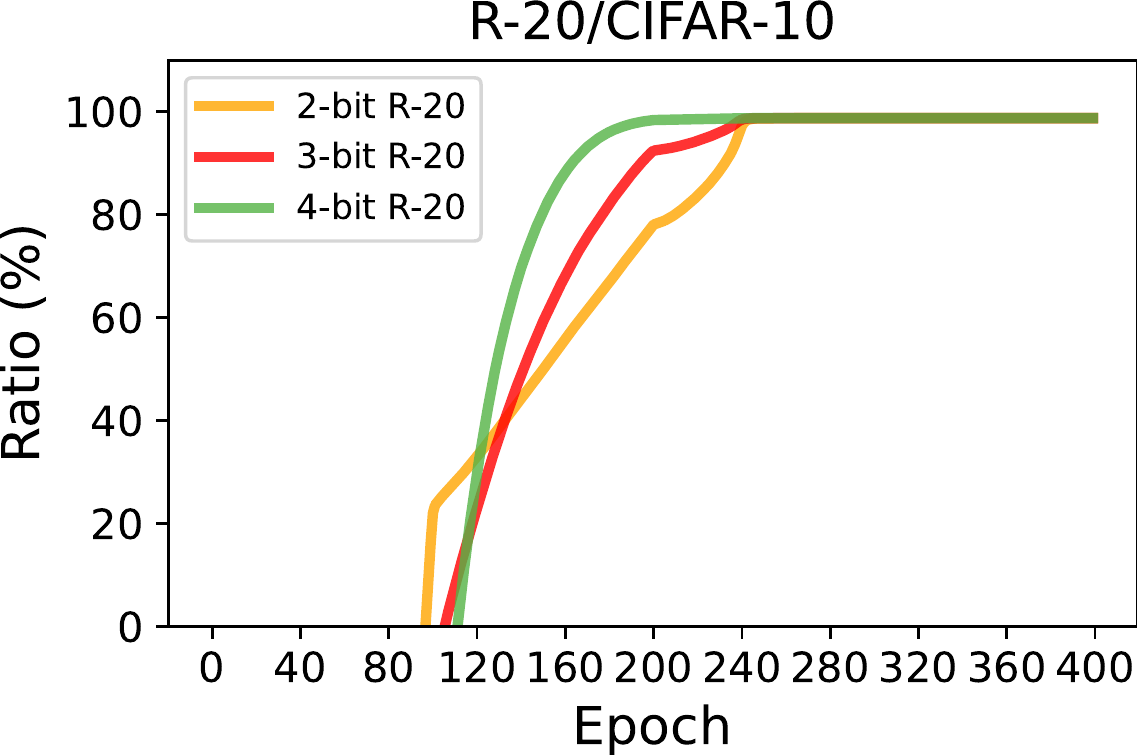}%
    \label{ill-sparsity:b}}
    \subfloat[]{
    \includegraphics[width=0.32\linewidth]{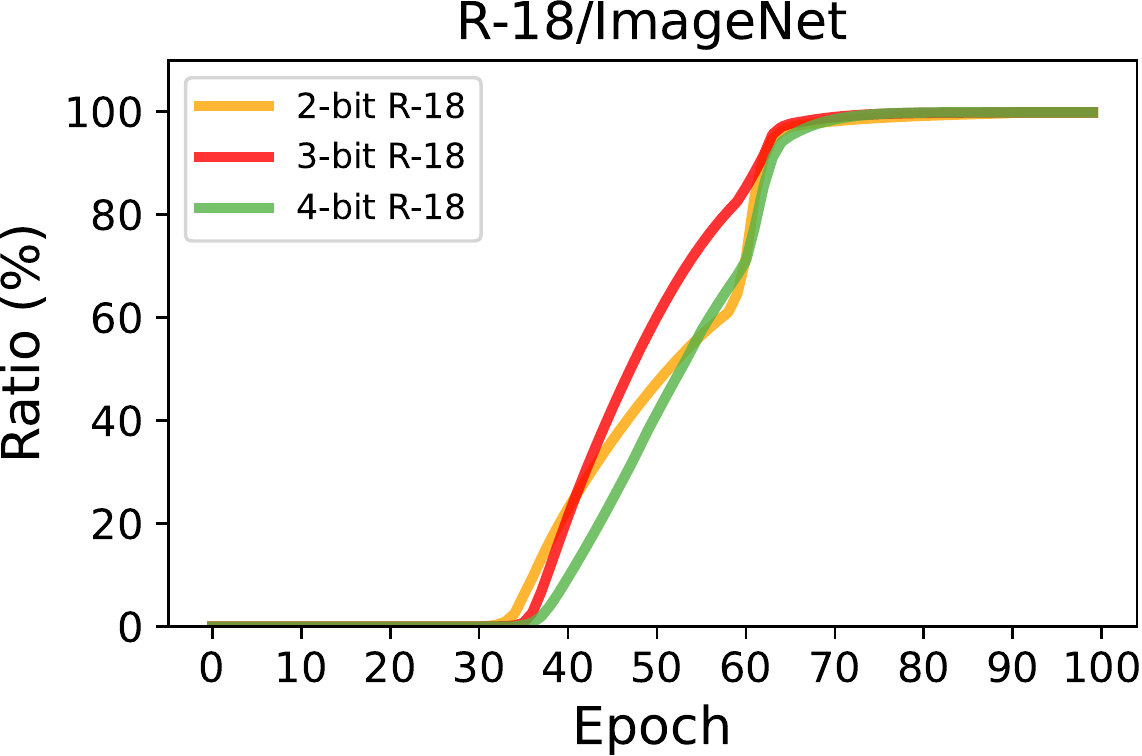}%
    \label{ill-sparsity:c}}
    \\
    \subfloat[]{
    \includegraphics[width=0.32\linewidth]{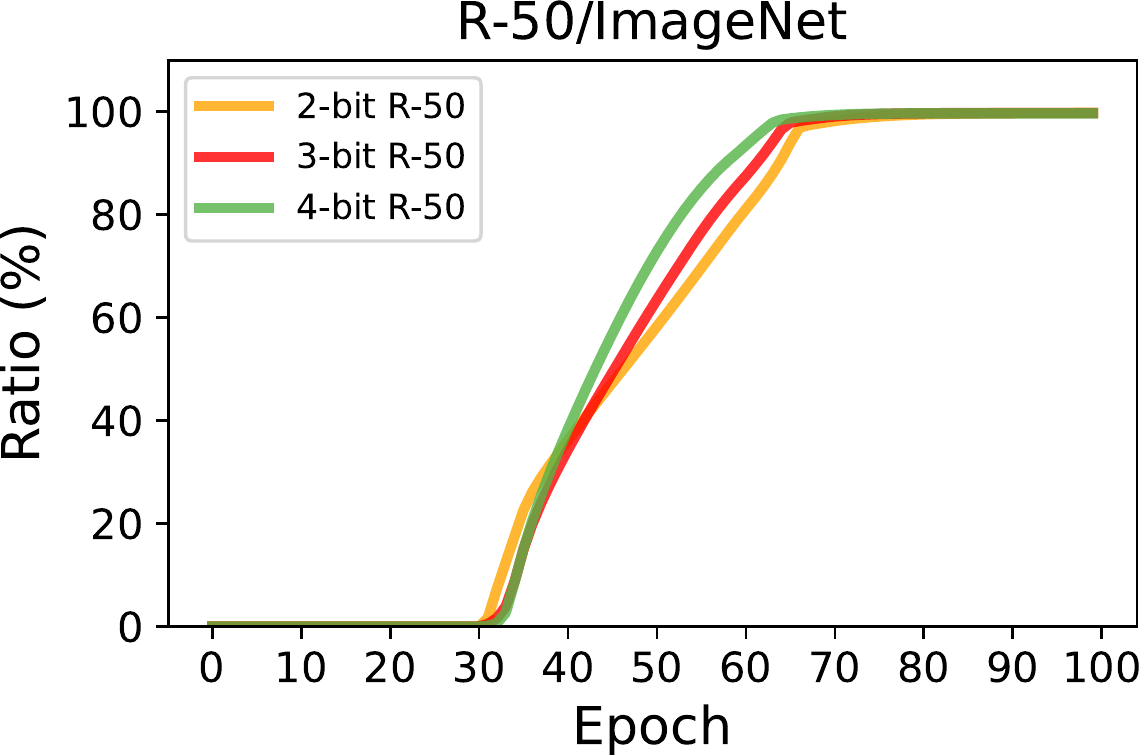}%
    \label{ill-sparsity:d}}
    \subfloat[]{
    \includegraphics[width=0.32\linewidth]{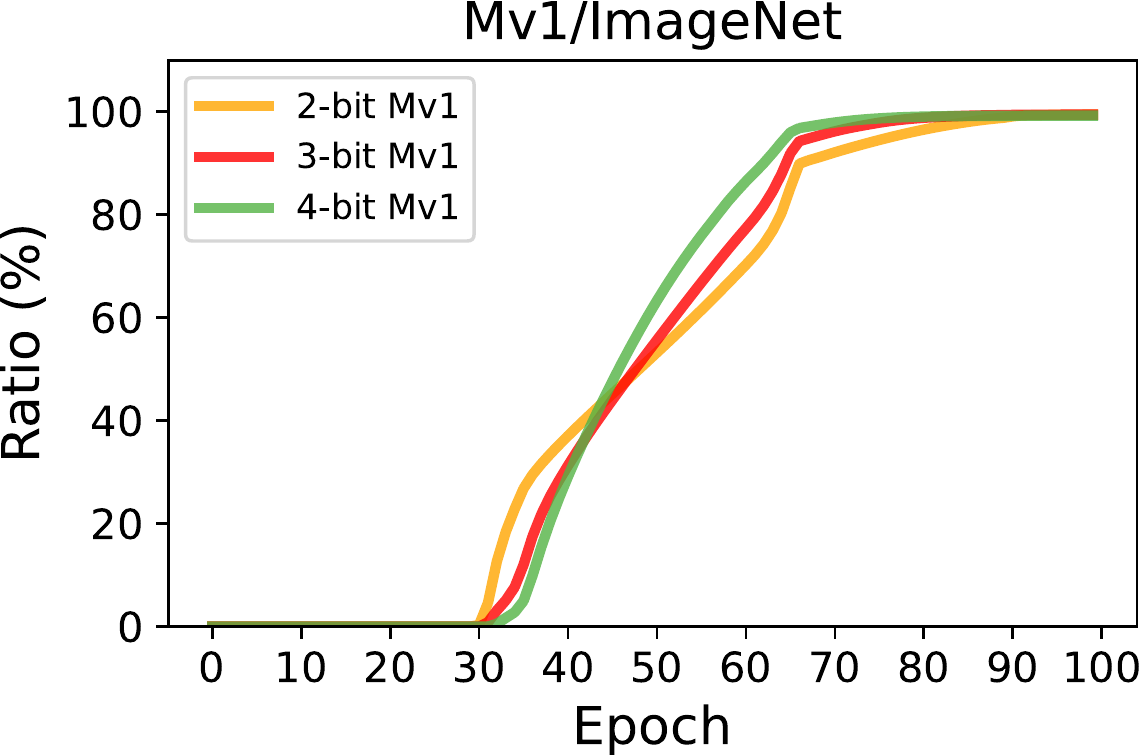}%
    \label{ill-sparsity:e}}
    \subfloat[]{
    \includegraphics[width=0.32\linewidth]{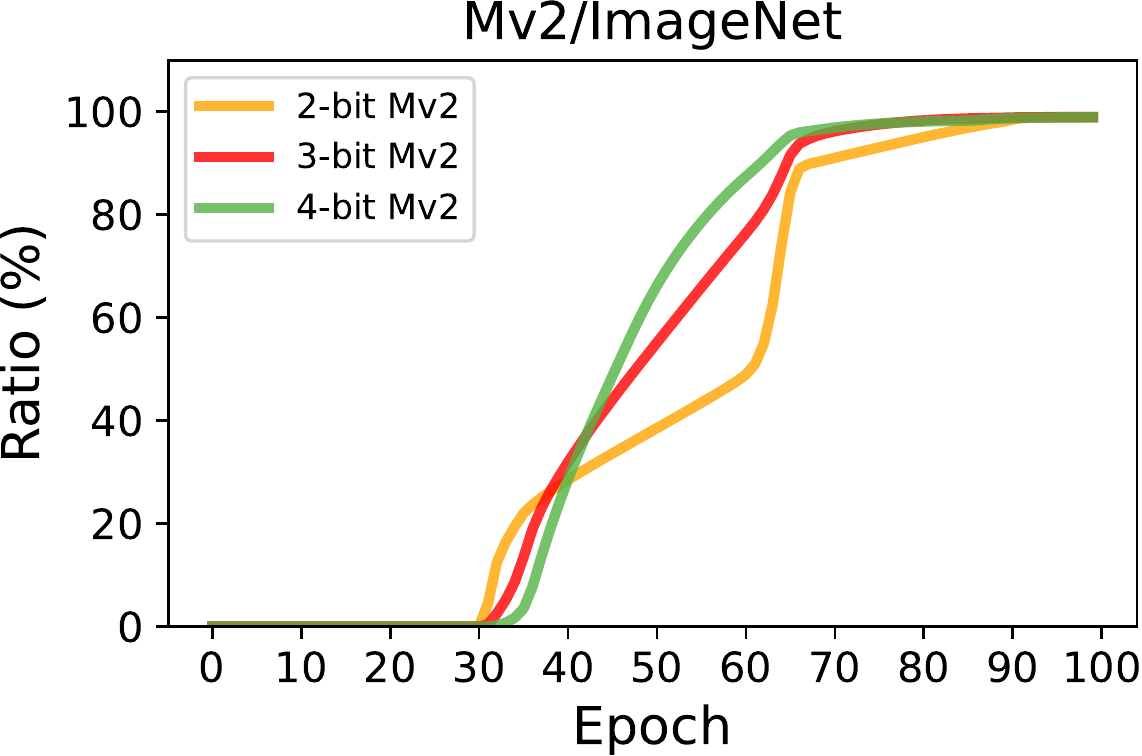}%
    \label{ill-sparsity:f}}
\caption{The weight gradient sparsity of quantized layers \emph{w.r.t.} training epochs. ``B-bit'' indicates the weights and activations are quantized to B-bit. ``R-20'', ``R-18'', ``R-50'', ``Mv1'', and ``Mv2'' indicate ResNet-20, ResNet-18, ResNet-50, MobileNetV1, and MobileNetV2, respectively.}
\label{ill-sparsity}
\end{figure*}

\subsubsection{ImageNet}

We then conduct the experiments on the challenging large-scale ImageNet. The results of ResNet-18, ResNet-50, MobileNetV1 and MobileNetV2 are presented in Tab.\,\ref{tab:ImageNet}.

\textbf{ResNet-18}. 
Results of the first part of Tab.\,\ref{tab:ImageNet} show that the proposed LTS achieve the comparable performance of or outperforms the baseline in terms of accuracy and efficiency across all bit-widths (2-, 3-, and 4-bit). Compared to the baseline, the proposed LTS achieved an average weight gradient sparsity of 49\%, 51\%, and 47\% for 2-, 3-, and 4-bits, respectively. This translates to a reduction in backward propagation FLOPs of 24.5\%, 25.5\%, and 23.5\%, respectively. For 3- and 4-bit, LTS only results in a sightly performance drop. Notably, the proposed LTS improves the baseline by 0.28\% for 2-bit quantization.

\textbf{ResNet-50}.
The results of ResNet-50 are presented in the second part of  Tab.\,\ref{tab:ImageNet}. It can be seen that the proposed LTS in general provides 50\% weight gradient sparsity, indicating the effectiveness of our LTS. Moreover, our LTS improves the compared baseline across all bit-widths. Specifically, on 2-, 3-, 4-bit, our LTS improves the performance by 0.74\%, 0.43\%, and 0.45\%, respectively. At the same time, our LTS is able to achieve 52\%, 52\%, and 53\% average of weight gradient sparsity, which gives 26\%, 26\%, and 26.5\% reduction in backward propagation FLOPs, respectively.

\textbf{MobileNetV1}. 
The results of MobileNetV1 in the third part of Tab.\,\ref{tab:ImageNet} demonstrate that the proposed LTS method is able to achieve significant weight gradient sparsity (ranging from 36\% to 52\%) for different bit-widths, indicating that the proposed method is effective in reducing the computational cost of QAT.
The proposed LTS achieves an average weight gradient sparsity of 50\%, 51\%, 52\%, 49\%, and 36\% for 2-, 3-, and 4-bit quantization, respectively, corresponding to a reduction in backward propagation FLOPs of 25\%, 25.5\%, and 26\%. 
Notably, the proposed method also improves the accuracy by 3.55\%, 1.59\%, and 0.50\% for 2-, 3-, and 4-bit quantization, respectively, compared to the baseline.

\textbf{MobileNetV2}. 
The fourth part of Tab.\,\ref{tab:ImageNet} provides the results of MobileNetV2. As can be seen, our LTS improves the performance of MobileNetV2 on 2-, 3-, and 4-bit. Specifically, LTS obtains accuracy gains 5.05\%, 0.82\%, and 0.60\% of on 2-, 3-, 4-bit MobileNetV2, respectively. The improvements are accompanied by a significant weight gradient sparsity, \emph{i.e.}, 46\%, 50\%, and 52\% for 2-, 3-, and 4-bit, respectively. Such sparsity gives 23\%, 25\%, and 26\% reduction in backward propagation FLOPs.

In summary, the proposed LTS either leads to improved performance or comparable performance compared with the baseline even though a high weight sparsity is achieved. Generally, on ImageNet, our LTS is able to obtain 46\% - 54\% average weight gradient sparsity and thus eliminate 23\% - 27\% FLOPs of backward propagation.
These findings suggest that the proposed LTS has the potential to be a valuable tool for efficient QAT.

\subsubsection{Discussion}
The performance improvements are most notable in low bit-widths and networks with efficient architecture. This can be attributed to the proposed LTS well solving the weight oscillations problem that is especially obvious for low bit-widths and efficient networks such as MobileNetV1~\cite{pmlr-v162-nagel22a}. 
The weight oscillations problem refers to the quantized weight periodically oscillating between two quantization levels, which leads to unstable training and inferior performance.
Although the original idea is to reduce training costs, our LTS subtly alleviates this problem by freezing weights during the training period and thus improves performance by a large margin.

\subsection{Comparisons With Random Frozen}

\begin{table}[ht]
\centering
\caption{Comparisons between different modes. Results are obtained from 2-bit ResNet-20 on CIFAR100 dataset.}
\label{tab:cifar-random}
\begin{tabular}{cccc}
\toprule[1.25pt]
Mode     & Accuracy. (\%) & Avg. of WGS & RD. FLOPs of BP   \\ \hline \hline
Baseline & 52.95     &  0\%                      &    0\%         \\
Random & 51.45     &  66\%                      &    33\%$\downarrow$         \\
\textbf{LTS} (Ours)    &   \textbf{53.76}    &   66\%               &       33\%$\downarrow$      \\\bottomrule[0.75pt]
\end{tabular}
\end{table}

In this subsection, we compare the proposed LTS with the random frozen mode, \emph{i.e.}, randomly freeze some weights. For a fair comparison, the ratio of randomly freeze weights is the same as the LTS.
Tab.\,\ref{tab:cifar-random} provides the results of 2-bit ResNet-20 on the CIFAR100 dataset. As can be observed, random freeze weight leads to a 1.50\% accuracy drop compared with the baseline, indicating the simple random freeze incurs performance loss. In contrast, our LTS improves the baseline performance by 0.81\%, clearly demonstrating the effectiveness of our method.

\subsection{Illustrations of Weight Gradient Sparsity}

We present the curve of the weight gradient sparsity of quantized weights \emph{w.r.t.} training epochs in Fig.\,\ref{ill-sparsity}.
It can be seen that the proposed LTS results in a gradual sparsity. The sparsity gradually increases until it reaches an extremely high magnitude.
Taking Fig.\,\ref{ill-sparsity:a} as an example, for ResNet-20 on CIFAR-100, the weight gradient sparsity increases rapidly after the warmup stage finishes. After epoch 240, the sparsity is very close to 100\%, which means most quantized weights are frozen.
As shown in Fig.\,\ref{ill-sparsity:b}, the results of ResNet-20 on CIFAR-10 are similar to CIFAR-100.
As for the large-scale ImageNet dataset, we observe that the sparsity starts to raise from epoch 30 to 33 depending on the bit-width. Then, the sparsity rapidly increases to a very high level. 
Concretely, Fig.\,\ref{ill-sparsity:c} provides the results of ResNet-18. Despite the raise point of sparsity increase being different slightly, our LTS still achieves near 80\% at epoch 60 in general. In particular, it can exceed 90\% at epoch 65.
Fig.\,\ref{ill-sparsity:c}-Fig.\,\ref{ill-sparsity:f} present the sparsity of ResNet-50, MobileNetV1, and MobileNetV2.
Despite the raise points of the sparsity of bit-widths being different, a high sparsity can be achieved around epoch 60 in general. 
For instance, in Fig.\,\ref{ill-sparsity:e}, the sparsity of 2-bit MobileNetV1 starts to increase quickly around epoch 30. While for 4-bit MobileNetV2, its raise point of sparsity increase is around epoch 33.
After the raise point, the sparsity raises quickly. For example, ResNet-50 of all bit-widths reaches 80\% sparsity around epoch 60.
Fig.\,\ref{ill-sparsity} demonstrates our LTS is able to create a high sparsity for the latter part of the training period.



\section{Conclusion}

In this paper, we challenge the contemporary experience in QAT that all quantized weights require updating throughout the entire training process.
We discover a straightforward-yet-valuable observation that a large portion of quantized weights, named as the partly scratch-off lottery ticket, converge to the optimal quantization level after a few training epochs.
Thus, weights within the ticket can be frozen, and thus their gradient calculations are zeroed out in the remaining training period to avoid meaningless updating.
We develop a heuristic method, dubbed lottery ticket scratcher (LTS), to effectively find the ticket. Specifically, a weight is frozen once the distance between it and its corresponding quantization level is lower than a threshold controlled by the proposed strategy. 
The introduced LTS is simple but extremely effective in accurately identifying the partly scratch-off lottery ticket and leading to a sparse weight gradient. 
Extensive experiments demonstrate that the proposed LTS generally achieves 50\%-70\% weight gradient sparsity and 25\%-35\% FLOPs reduction of the backward pass, while still exhibiting comparable or even better performance than the compared baseline.

\bibliographystyle{IEEEtran}
\bibliography{egbib}

\begin{thebibliography}{10}
\providecommand{\url}[1]{#1}
\csname url@samestyle\endcsname
\providecommand{\newblock}{\relax}
\providecommand{\bibinfo}[2]{#2}
\providecommand{\BIBentrySTDinterwordspacing}{\spaceskip=0pt\relax}
\providecommand{\BIBentryALTinterwordstretchfactor}{4}
\providecommand{\BIBentryALTinterwordspacing}{\spaceskip=\fontdimen2\font plus
\BIBentryALTinterwordstretchfactor\fontdimen3\font minus
  \fontdimen4\font\relax}
\providecommand{\BIBforeignlanguage}[2]{{%
\expandafter\ifx\csname l@#1\endcsname\relax
\typeout{** WARNING: IEEEtran.bst: No hyphenation pattern has been}%
\typeout{** loaded for the language `#1'. Using the pattern for}%
\typeout{** the default language instead.}%
\else
\language=\csname l@#1\endcsname
\fi
#2}}
\providecommand{\BIBdecl}{\relax}
\BIBdecl

\bibitem{han2015learning}
S.~Han, J.~Pool, J.~Tran, W.~J. Dally \emph{et~al.}, ``Learning both weights
  and connections for efficient neural network,'' in \emph{Proceedings of the
  Advances in Neural Information Processing Systems (NeurIPS)}, 2015, pp.
  1135--1143.

\bibitem{whitepaper}
R.~Krishnamoorthi, ``Quantizing deep convolutional networks for efficient
  inference: A whitepaper,'' \emph{arXiv preprint arXiv:1806.08342}, 2018.

\bibitem{hinton2015distilling}
G.~Hinton, O.~Vinyals, and J.~Dean, ``Distilling the knowledge in a neural
  network,'' \emph{arXiv preprint arXiv:1503.02531}, 2015.

\bibitem{lin2020hrank}
M.~Lin, R.~Ji, Y.~Wang, Y.~Zhang, B.~Zhang, Y.~Tian, and L.~Shao, ``Hrank:
  Filter pruning using high-rank feature map,'' in \emph{Proceedings of the
  IEEE/CVF Conference on Computer Vision and Pattern Recognition (CVPR)}, 2020,
  pp. 1529--1538.

\bibitem{ACIQ}
R.~Banner, Y.~Nahshan, D.~Soudry \emph{et~al.}, ``Post training 4-bit
  quantization of convolutional networks for rapid-deployment,'' in
  \emph{Proceedings of the Advances in Neural Information Processing Systems
  (NeurIPS)}, 2019, pp. 7950--7958.

\bibitem{lin2020rotated}
M.~Lin, R.~Ji, Z.~Xu, B.~Zhang, Y.~Wang, Y.~Wu, F.~Huang, and C.-W. Lin,
  ``Rotated binary neural network,'' in \emph{Proceedings of the Advances in
  Neural Information Processing Systems (NeurIPS)}, 2020, pp. 7474--7485.

\bibitem{zhong2022intraq}
Y.~Zhong, M.~Lin, G.~Nan, J.~Liu, B.~Zhang, Y.~Tian, and R.~Ji, ``Intraq:
  Learning synthetic images with intra-class heterogeneity for zero-shot
  network quantization,'' in \emph{Proceedings of the IEEE/CVF Conference on
  Computer Vision and Pattern Recognition (CVPR)}, 2022, pp. 12\,339--12\,348.

\bibitem{bulat2020HighCapacity}
A.~Bulat, B.~Martinez, and G.~Tzimiropoulos, ``High-capacity expert binary
  networks,'' in \emph{Proceedings of the International Conference on Learning
  Representations (ICLR)}, 2020.

\bibitem{gong2019differentiable}
R.~Gong, X.~Liu, S.~Jiang, T.~Li, P.~Hu, J.~Lin, F.~Yu, and J.~Yan,
  ``Differentiable soft quantization: Bridging full-precision and low-bit
  neural networks,'' in \emph{Proceedings of the IEEE/CVF Conference on
  Computer Vision and Pattern Recognition (CVPR)}, 2019, pp. 4852--4861.

\bibitem{gao2022clusterq}
Y.~Gao, Z.~Zhang, R.~Hong, H.~Zhang, J.~Fan, S.~Yan, and M.~Wang, ``Clusterq:
  Semantic feature distribution alignment for data-free quantization,''
  \emph{arXiv preprint arXiv:2205.00179}, 2022.

\bibitem{IntegerOnly}
B.~Jacob, S.~Kligys, B.~Chen, M.~Zhu, M.~Tang, A.~Howard, H.~Adam, and
  D.~Kalenichenko, ``Quantization and training of neural networks for efficient
  integer-arithmetic-only inference,'' in \emph{Proceedings of the IEEE/CVF
  Conference on Computer Vision and Pattern Recognition (CVPR)}, 2018, pp.
  2704--2713.

\bibitem{he2016deep}
K.~He, X.~Zhang, S.~Ren, and J.~Sun, ``Deep residual learning for image
  recognition,'' in \emph{Proceedings of the IEEE/CVF Conference on Computer
  Vision and Pattern Recognition (CVPR)}, 2016, pp. 770--778.

\bibitem{cifar}
A.~Krizhevsky, ``Learning multiple layers of features from tiny images,''
  \emph{University of Toronto}, 2009.

\bibitem{frankle2018lottery}
J.~Frankle and M.~Carbin, ``The lottery ticket hypothesis: Finding sparse,
  trainable neural networks,'' in \emph{Proceedings of the International
  Conference on Learning Representations (ICLR)}, 2019.

\bibitem{bellec2018deep}
G.~Bellec, D.~Kappel, W.~Maass, and R.~Legenstein, ``Deep rewiring: Training
  very sparse deep networks,'' in \emph{Proceedings of the International
  Conference on Learning Representations (ICLR)}, 2018.

\bibitem{mostafa2019parameter}
H.~Mostafa and X.~Wang, ``Parameter efficient training of deep convolutional
  neural networks by dynamic sparse reparameterization,'' in \emph{Proceedings
  of the International Conference on Machine Learning (ICML)}.\hskip 1em plus
  0.5em minus 0.4em\relax PMLR, 2019, pp. 4646--4655.

\bibitem{liu2021sparse}
S.~Liu, T.~Chen, X.~Chen, Z.~Atashgahi, L.~Yin, H.~Kou, L.~Shen,
  M.~Pechenizkiy, Z.~Wang, and D.~C. Mocanu, ``Sparse training via boosting
  pruning plasticity with neuroregeneration,'' in \emph{Proceedings of the
  Advances in Neural Information Processing Systems (NeurIPS)}, 2021, pp.
  9908--9922.

\bibitem{pmlr-v162-nagel22a}
M.~Nagel, M.~Fournarakis, Y.~Bondarenko, and T.~Blankevoort, ``Overcoming
  oscillations in quantization-aware training,'' in \emph{Proceedings of the
  International Conference on Machine Learning (ICML)}, 2022, pp.
  16\,318--16\,330.

\bibitem{LSQ}
S.~K. Esser, J.~L. McKinstry, D.~Bablani, R.~Appuswamy, and D.~S. Modha,
  ``Learned step size quantization,'' in \emph{Proceedings of the International
  Conference on Learning Representations (ICLR)}, 2020.

\bibitem{APoT}
Y.~Li, X.~Dong, and W.~Wang, ``Additive powers-of-two quantization: An
  efficient non-uniform discretization for neural networks,'' in
  \emph{Proceedings of the International Conference on Learning Representations
  (ICLR)}, 2020.

\bibitem{jung2019learning}
S.~Jung, C.~Son, S.~Lee, J.~Son, J.-J. Han, Y.~Kwak, S.~J. Hwang, and C.~Choi,
  ``Learning to quantize deep networks by optimizing quantization intervals
  with task loss,'' in \emph{Proceedings of the IEEE/CVF Conference on Computer
  Vision and Pattern Recognition (CVPR)}, 2019, pp. 4350--4359.

\bibitem{liu2022nonuniform}
Z.~Liu, K.-T. Cheng, D.~Huang, E.~P. Xing, and Z.~Shen, ``Nonuniform-to-uniform
  quantization: Towards accurate quantization via generalized straight-through
  estimation,'' in \emph{Proceedings of the IEEE/CVF Conference on Computer
  Vision and Pattern Recognition (CVPR)}, 2022, pp. 4942--4952.

\bibitem{zhang2021differentiable}
Z.~Zhang, W.~Shao, J.~Gu, X.~Wang, and P.~Luo, ``Differentiable dynamic
  quantization with mixed precision and adaptive resolution,'' in
  \emph{Proceedings of the International Conference on Machine Learning
  (ICML)}.\hskip 1em plus 0.5em minus 0.4em\relax PMLR, 2021, pp.
  12\,546--12\,556.

\bibitem{zhou2016incremental}
A.~Zhou, A.~Yao, Y.~Guo, L.~Xu, and Y.~Chen, ``Incremental network
  quantization: Towards lossless cnns with low-precision weights,'' in
  \emph{Proceedings of the International Conference on Learning Representations
  (ICLR)}, 2017.

\bibitem{zhuang2020training}
B.~Zhuang, L.~Liu, M.~Tan, C.~Shen, and I.~Reid, ``Training quantized neural
  networks with a full-precision auxiliary module,'' in \emph{Proceedings of
  the IEEE/CVF Conference on Computer Vision and Pattern Recognition (CVPR)},
  2020, pp. 1488--1497.

\bibitem{lee2021network}
J.~Lee, D.~Kim, and B.~Ham, ``Network quantization with element-wise gradient
  scaling,'' in \emph{Proceedings of the IEEE/CVF Conference on Computer Vision
  and Pattern Recognition (CVPR)}, 2021, pp. 6448--6457.

\bibitem{kim2021distance}
D.~Kim, J.~Lee, and B.~Ham, ``Distance-aware quantization,'' in
  \emph{Proceedings of the IEEE/CVF International Conference on Computer Vision
  (ICCV)}, 2021, pp. 5271--5280.

\bibitem{lee2021cluster}
J.~H. Lee, J.~Yun, S.~J. Hwang, and E.~Yang, ``Cluster-promoting quantization
  with bit-drop for minimizing network quantization loss,'' in
  \emph{Proceedings of the IEEE/CVF International Conference on Computer Vision
  (ICCV)}, 2021, pp. 5370--5379.

\bibitem{han2021improving}
T.~Han, D.~Li, J.~Liu, L.~Tian, and Y.~Shan, ``Improving low-precision network
  quantization via bin regularization,'' in \emph{Proceedings of the IEEE/CVF
  International Conference on Computer Vision (ICCV)}, 2021, pp. 5261--5270.

\bibitem{dong2019hawq}
Z.~Dong, Z.~Yao, A.~Gholami, M.~W. Mahoney, and K.~Keutzer, ``Hawq: Hessian
  aware quantization of neural networks with mixed-precision,'' in
  \emph{Proceedings of the IEEE/CVF International Conference on Computer Vision
  (ICCV)}, 2019, pp. 293--302.

\bibitem{yang2021fracbits}
L.~Yang and Q.~Jin, ``Fracbits: Mixed precision quantization via fractional
  bit-widths,'' in \emph{Proceedings of the AAAI Conference on Artificial
  Intelligence (AAAI)}, vol.~35, 2021, pp. 10\,612--10\,620.

\bibitem{chen2021towards}
W.~Chen, P.~Wang, and J.~Cheng, ``Towards mixed-precision quantization of
  neural networks via constrained optimization,'' in \emph{Proceedings of the
  IEEE/CVF International Conference on Computer Vision (ICCV)}, 2021, pp.
  5350--5359.

\bibitem{zhou2019deconstructing}
H.~Zhou, J.~Lan, R.~Liu, and J.~Yosinski, ``Deconstructing lottery tickets:
  zeros, signs, and the supermask,'' in \emph{Advances in Neural Information
  Processing Systems (NeurIPS)}, 2019, pp. 3597--3607.

\bibitem{ramanujan2020s}
V.~Ramanujan, M.~Wortsman, A.~Kembhavi, A.~Farhadi, and M.~Rastegari, ``What's
  hidden in a randomly weighted neural network?'' in \emph{IEEE Conference on
  Computer Vision and Pattern Recognition (CVPR)}, 2020, pp. 11\,893--11\,902.

\bibitem{zhang2021lottery}
Y.~Zhang, M.~Lin, F.~Chao, Y.~Wang, K.~Li, Y.~Shen, Y.~Wu, and R.~Ji, ``Lottery
  jackpots exist in pre-trained models,'' \emph{arXiv preprint
  arXiv:2104.08700}, 2021.

\bibitem{wang2020picking}
C.~Wang, G.~Zhang, and R.~Grosse, ``Picking winning tickets before training by
  preserving gradient flow,'' in \emph{International Conference on Learning
  Representations (ICLR)}, 2020.

\bibitem{pensia2020optimal}
A.~Pensia, S.~Rajput, A.~Nagle, H.~Vishwakarma, and D.~Papailiopoulos,
  ``Optimal lottery tickets via subsetsum: Logarithmic over-parameterization is
  sufficient,'' in \emph{Advances in Neural Information Processing Systems
  (NeurIPS)}, 2020, pp. 2599--2610.

\bibitem{goyal2017accurate}
P.~Goyal, P.~Doll{\'a}r, R.~Girshick, P.~Noordhuis, L.~Wesolowski, A.~Kyrola,
  A.~Tulloch, Y.~Jia, and K.~He, ``Accurate, large minibatch sgd: Training
  imagenet in 1 hour,'' \emph{arXiv preprint arXiv:1706.02677}, 2017.

\bibitem{jia2018highly}
X.~Jia, S.~Song, W.~He, Y.~Wang, H.~Rong, F.~Zhou, L.~Xie, Z.~Guo, Y.~Yang,
  L.~Yu \emph{et~al.}, ``Highly scalable deep learning training system with
  mixed-precision: Training imagenet in four minutes,'' \emph{arXiv preprint
  arXiv:1807.11205}, 2018.

\bibitem{you2018imagenet}
Y.~You, Z.~Zhang, C.-J. Hsieh, J.~Demmel, and K.~Keutzer, ``Imagenet training
  in minutes,'' in \emph{Proceedings of the 47th International Conference on
  Parallel Processing}, 2018, pp. 1--10.

\bibitem{akiba2017extremely}
T.~Akiba, S.~Suzuki, and K.~Fukuda, ``Extremely large minibatch sgd: Training
  resnet-50 on imagenet in 15 minutes,'' \emph{arXiv preprint
  arXiv:1711.04325}, 2017.

\bibitem{sun2017meprop}
X.~Sun, X.~Ren, S.~Ma, and H.~Wang, ``meprop: Sparsified back propagation for
  accelerated deep learning with reduced overfitting,'' in \emph{Proceedings of
  the International Conference on Machine Learning (ICML)}.\hskip 1em plus
  0.5em minus 0.4em\relax PMLR, 2017, pp. 3299--3308.

\bibitem{goli2020resprop}
N.~Goli and T.~M. Aamodt, ``Resprop: Reuse sparsified backpropagation,'' in
  \emph{Proceedings of the IEEE/CVF Conference on Computer Vision and Pattern
  Recognition (CVPR)}, 2020, pp. 1548--1558.

\bibitem{evci2020rigging}
U.~Evci, T.~Gale, J.~Menick, P.~S. Castro, and E.~Elsen, ``Rigging the lottery:
  Making all tickets winners,'' in \emph{Proceedings of the International
  Conference on Machine Learning (ICML)}.\hskip 1em plus 0.5em minus
  0.4em\relax PMLR, 2020, pp. 2943--2952.

\bibitem{raihan2020sparse}
M.~A. Raihan and T.~Aamodt, ``Sparse weight activation training,'' in
  \emph{Proceedings of the Advances in Neural Information Processing Systems
  (NeurIPS)}, vol.~33, 2020, pp. 15\,625--15\,638.

\bibitem{evci2021gradmax}
U.~Evci, B.~van Merrienboer, T.~Unterthiner, F.~Pedregosa, and M.~Vladymyrov,
  ``Gradmax: Growing neural networks using gradient information,'' in
  \emph{Proceedings of the International Conference on Learning Representations
  (ICLR)}, 2022.

\bibitem{liu2021we}
S.~Liu, L.~Yin, D.~C. Mocanu, and M.~Pechenizkiy, ``Do we actually need dense
  over-parameterization? in-time over-parameterization in sparse training,'' in
  \emph{Proceedings of the International Conference on Machine Learning
  (ICML)}.\hskip 1em plus 0.5em minus 0.4em\relax PMLR, 2021, pp. 6989--7000.

\bibitem{banner2018scalable}
R.~Banner, I.~Hubara, E.~Hoffer, and D.~Soudry, ``Scalable methods for 8-bit
  training of neural networks,'' in \emph{Proceedings of the Advances in Neural
  Information Processing Systems (NeurIPS)}, 2018, pp. 5151--5159.

\bibitem{zhu2020towards}
F.~Zhu, R.~Gong, F.~Yu, X.~Liu, Y.~Wang, Z.~Li, X.~Yang, and J.~Yan, ``Towards
  unified int8 training for convolutional neural network,'' in
  \emph{Proceedings of the IEEE/CVF Conference on Computer Vision and Pattern
  Recognition (CVPR)}, 2020, pp. 1969--1979.

\bibitem{zhao2021distribution}
K.~Zhao, S.~Huang, P.~Pan, Y.~Li, Y.~Zhang, Z.~Gu, and Y.~Xu, ``Distribution
  adaptive int8 quantization for training cnns,'' in \emph{Proceedings of the
  AAAI Conference on Artificial Intelligence (AAAI)}, vol.~35, 2021, pp.
  3483--3491.

\bibitem{lee2021toward}
S.~Lee, J.~Park, and D.~Jeon, ``Toward efficient low-precision training: Data
  format optimization and hysteresis quantization,'' in \emph{Proceedings of
  the International Conference on Learning Representations (ICLR)}, 2022.

\bibitem{chmiel2020neural}
B.~Chmiel, L.~Ben-Uri, M.~Shkolnik, E.~Hoffer, R.~Banner, and D.~Soudry,
  ``Neural gradients are near-lognormal: improved quantized and sparse
  training,'' in \emph{Proceedings of the International Conference on Learning
  Representations (ICLR)}, 2021.

\bibitem{hu2022palquant}
Q.~Hu, G.~Li, Q.~Wu, and J.~Cheng, ``Palquant: Accelerating high-precision
  networks on low-precision accelerators,'' in \emph{Proceedings of the
  European Conference on Computer Vision (ECCV)}.\hskip 1em plus 0.5em minus
  0.4em\relax Springer, 2022, pp. 312--327.

\bibitem{russakovsky2015imagenet}
O.~Russakovsky, J.~Deng, H.~Su, J.~Krause, S.~Satheesh, S.~Ma, Z.~Huang,
  A.~Karpathy, A.~Khosla, M.~Bernstein \emph{et~al.}, ``Imagenet large scale
  visual recognition challenge,'' \emph{International Journal of Computer
  Vision (IJCV)}, vol. 115, pp. 211--252, 2015.

\bibitem{howard2017mobilenets}
A.~G. Howard, M.~Zhu, B.~Chen, D.~Kalenichenko, W.~Wang, T.~Weyand,
  M.~Andreetto, and H.~Adam, ``Mobilenets: Efficient convolutional neural
  networks for mobile vision applications,'' \emph{arXiv preprint
  arXiv:1704.04861}, 2017.

\bibitem{sandler2018mobilenetv2}
M.~Sandler, A.~Howard, M.~Zhu, A.~Zhmoginov, and L.-C. Chen, ``Mobilenetv2:
  Inverted residuals and linear bottlenecks,'' in \emph{Proceedings of the
  IEEE/CVF Conference on Computer Vision and Pattern Recognition (CVPR)}, 2018,
  pp. 4510--4520.

\bibitem{paszke2019pytorch}
A.~Paszke, S.~Gross, F.~Massa, A.~Lerer, J.~Bradbury, G.~Chanan, T.~Killeen,
  Z.~Lin, N.~Gimelshein, L.~Antiga \emph{et~al.}, ``Pytorch: An imperative
  style, high-performance deep learning library,'' in \emph{Proceedings of the
  Advances in Neural Information Processing Systems (NeurIPS)}, 2019, pp.
  8026--8037.

\bibitem{courbariaux2016binarized}
M.~Courbariaux, I.~Hubara, D.~Soudry, R.~El-Yaniv, and Y.~Bengio, ``Binarized
  neural networks: Training deep neural networks with weights and activations
  constrained to+ 1 or-1,'' \emph{arXiv preprint arXiv:1602.02830}, 2016.

\end{thebibliography}

\end{document}